\pdfoutput=1

\documentclass[11pt]{article}

\usepackage[preprint]{acl}

\usepackage{times}
\usepackage{latexsym}
\usepackage{hyperref}
\usepackage[T1]{fontenc}

\usepackage[utf8]{inputenc}

\usepackage{microtype}
\usepackage{inconsolata}
\usepackage{makecell}
\usepackage{amsmath} 
\usepackage{amssymb}
\usepackage{graphicx}
\usepackage{booktabs,tabularx}
\usepackage{adjustbox}
\usepackage[table]{xcolor} 
\usepackage{enumitem}
\usepackage{todonotes}

\usepackage{listings}
\usepackage{xcolor}
\usepackage{tcolorbox}
\tcbuselibrary{listings,breakable,skins}
\usepackage{geometry}

\newtcolorbox{guidelinebox}{
    enhanced,
    breakable,
    colback=blue!4,
    colframe=blue!55!black,
    boxrule=0.4pt,
    arc=2pt,
    left=2pt, right=2pt, top=2pt, bottom=2pt,
    before skip=4pt, after skip=4pt,
}

\usepackage{subcaption}
\usepackage{multirow}
\usepackage{pifont}

\newcommand{\affilsup}[1]{\rlap{\textsuperscript{\normalfont#1}}}

\title{What Are We Measuring in NLG? A Meta-Analysis of Evaluation Trends 2020-2025}

\author{
    Jing Yang\affilsup{1,2}
    \qquad
    Nils Feldhus\affilsup{7,3,\footnotemark[1]}
    \qquad
    Salar Mohtaj\affilsup{3}
    \qquad \\
    \textbf{Leonhard Hennig}\affilsup{3} 
    \qquad 
    \textbf{Qianli Wang}\affilsup{1}
    \qquad 
    \textbf{Eleni Metheniti}\affilsup{4}
    \qquad \\
    \textbf{Sherzod Hakimov}\affilsup{5}
    \qquad 
    \textbf{Charlott Jakob}\affilsup{1,3}
    \qquad 
    \textbf{Veronika Solopova}\affilsup{1}
    \qquad \\
    \textbf{Konrad Rieck}\affilsup{1,2}
     \qquad
    \textbf{David Schlangen}\affilsup{3,5}
    \qquad
    \textbf{Sebastian M\"oller}\affilsup{1,2,3}
    \qquad
    \textbf{Vera Schmitt}\affilsup{1,3,6,8}
    \\
    $^1$Technische Universit\"at Berlin \\
    $^2$BIFOLD – Berlin Institute for the Foundations of Learning and Data \\
    $^3$German Research Center for Artificial Intelligence (DFKI) \\
    $^4$Joint Research Center \qquad 
    $^5$University of Potsdam \qquad 
    $^6$CERTAIN\\
    $^7$University of Groningen
    \qquad 
    $^8$Johannes Gutenberg University Mainz \\
}

\begin{document}
\maketitle
\renewcommand{\thefootnote}{\fnsymbol{footnote}}
\footnotetext[1]{Work done at TU Berlin and BIFOLD.}
\renewcommand*{\thefootnote}{\arabic{footnote}}

\begin{abstract}
 
As Natural Language Generation (NLG) dominates modern NLP, scalable evaluation remains a critical bottleneck. Consequently, LLM-as-a-judge (LaaJ) adoption has accelerated rapidly, appearing in more papers than human evaluation in 2025. This pivotal shift motivates a critical analysis of current evaluation practices.  
Overcoming the limits of rigid keyword filtering and manual review, we employ a multi-LLM information extraction pipeline to gather structured metadata from 14,171 papers across four major NLP conferences (2020--2025). Analyzing 3,334 filtered NLG papers, we identify three systemic challenges. (1) Metric inertia: despite the shift toward open-ended generation, legacy lexical metrics (BLEU, ROUGE) persist as primary indicators, typically used alongside rather than replaced by semantic alternatives. (2) Metric-criteria mapping problem: our paper-level co-occurrence data reveals that general-purpose automatic metrics are applied as broad proxies for quality, without specifying which dimension of text generation they are intended to evaluate. (3) Validation gap: LaaJ has grown rapidly without commensurate human validation (fewer than 8\% of papers). Crucially, while LaaJ correlates with aggregate quality, alignment collapses on fine-grained criteria like fluency. To address these gaps, we distill our findings into a minimal Evaluation Checklist to guide metric selection, construct validity, and LaaJ deployment.\footnote{Code and data are available at: \url{https://github.com/jingyng/NLG-Evaluation-Trends}.}

\end{abstract}

\section{Introduction}

\begin{figure}[!t]
    \centering
    \includegraphics[width=\columnwidth]{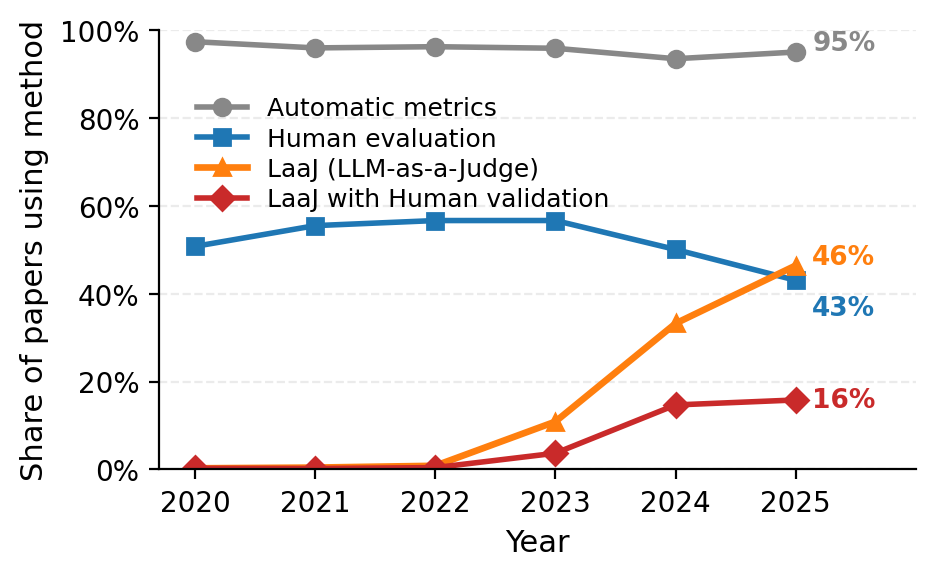}
    \caption{Evaluation method adoption across the 3,334 Top-30 NLG-task papers (2020--2025). Automatic metrics remain ubiquitous ($\sim$95\%). LaaJ adoption surged from less than 1\% before 2023 to 46\% in 2025, surpassing human evaluation, which declined from 51\% to 43\%.  Yet only 16\% of papers in 2025 explicitly validate LaaJ against human evaluation, indicating a validation~gap.}
    \label{fig:eval_method_adoption}
\end{figure}

Over the past few years, significant developments have taken place in NLG. While the previous generation of models focused on a limited range of tasks like summarization and translation, the dominance of transformer-based Large Language Models (LLMs) has enabled substantially more flexible applications, ranging from poetry to medical report generation.
This shift has reshaped evaluation paradigms and introduced new challenges, such as dealing with confabulations, ensuring factual consistency, and mitigating biases \cite{gehrmann-2023-repairing-the-cracked-foundation}. Consequently, metrics have evolved beyond simple lexical overlap. Newer semantic-aware metrics have emerged, including BERTScore \cite{zhangbertscore} and BLEURT \cite{sellam-etal-2020-bleurt}, and most notably, the surge of reference-free, prompt-based LaaJ methods \cite{gao-etal-2025-llmeval, gu2025surveyllmasajudge, li2024llmsasjudgescomprehensivesurveyllmbased}. 
Figure~\ref{fig:eval_method_adoption} shows that LaaJ adoption surged from less than 1\% of papers before 2023 to 46\% in 2025, surpassing human evaluation, which has remained flat-to-declining. Yet only 16\% of 2025 papers explicitly validate LaaJ against human evaluation, exposing a validation gap that we quantify throughout this~work.

Motivated by the need to better understand how text quality evaluation is conducted in NLG research, we employ LLMs as research tools \cite{liao-2025-llms-as-research-tools} to analyze the literature in major NLP conferences. This approach is critical to bridge the gap between scale and depth: it allows us to filter papers with a semantic accuracy that keyword searches lack, and extract structured metadata at a volume impossible for manual review. Our contributions are summarized as follows:

\begin{itemize}[nosep,labelsep=0.2em,align=parleft,topsep=0pt,partopsep=0pt]
    \item We present the largest study of NLG evaluation to date, applying multi-LLM extraction to 14,171 papers, and analyzing 3,334 NLG papers. Our human validation of the results reveals strong agreement between human and LLM annotators.
    \item We identify three persistent challenges in NLG evaluation: metric inertia, where lexical metrics persist alongside semantic alternatives; a metric–criteria mapping problem, where general-purpose metrics appear broadly across tasks rather than systematically pairing with specific evaluation constructs; and a validation gap where LaaJ has expanded without proportional human validation.
    \item We provide an evaluation checklist grounded in our findings, intended as concrete starting points for more rigorous NLG evaluation.

\end{itemize}

\section{Related Work}

\paragraph{LLMs for Annotation of Research Papers.}
 
\citet{tan-etal-2024-large} point out that LLMs can reliably annotate instructions and responses, even for specific domains, with human-level quality. 
A few prior works have dealt with the automated processing of scientific literature \cite{agarwal-2025-litllms, du-2024-llms-assist-nlp-researchers, scherbakov-2025-llms-as-tools-in-lit-reviews}. Existing survey papers making use of LLMs as annotators typically rely on prompts that assess whether a paper is relevant to the target topic \cite{alabi-2025-charting-landscape-african-nlp, alyafeai-2025-mole} or simply collect metadata like citation counts \cite{bernasconi-2025-integrated-survey-classification}.
Unlike previous work, we provide the first systematic, large-scale annotation of research papers, not only for paper relevance filtering but also for quantifying research trends and providing critical overviews.

\paragraph{Surveys in NLG evaluation.}

\citet{sai-2022-evaluation-metrics-nlg-systems} provide a taxonomy of NLG metrics and highlight that traditional reference-based metrics often correlate poorly with human judgment and fail to capture nuances like factual correctness.
With the rise of LLMs, researchers have begun using them as evaluators (i.e., LaaJ) to assess generated text quality. This marks a new direction in NLG evaluation that earlier surveys did not address~\citep{celikyilmaz-2021-evaluation-text-generation, sai-2022-evaluation-metrics-nlg-systems}. 
Recent surveys explicitly taxonomize these methods and highlight reliability challenges \cite{gao-etal-2025-llmeval, gu2025surveyllmasajudge}.
While LaaJ can replicate human judgments on certain criteria, it exhibits substantial variability across tasks \cite{bavaresco-etal-2025-llms} and often struggles with generalization \cite{huang-etal-2025-empirical} and reliability \cite{hu-2024-are-llm-based-evaluators-confusing-nlg-quality-criteria, wang-2024-llms-are-not-fair-evaluators, lee-2025-evaluating-consistency-llm-evaluators, wang-2025-dhp}.
While humans remain the most reliable and trusted evaluators, several meta-analyses have highlighted the inconsistency in human evaluation protocols for NLG and the need for standardized guidelines~\cite{van-der-lee-etal-2019-best, howcroft-etal-2020-twenty, celikyilmaz-2021-evaluation-text-generation}. 
While recent manual surveys identified metric shortcomings across 110 papers \cite{schmidtova-2024-automatic-metrics-in-nlg}, we use multi-LLM extraction to massively scale annotations to thousands of papers across a six-year span.

\begin{figure*}[t]
    \centering
    \resizebox{\textwidth}{!}{%
	    \includegraphics{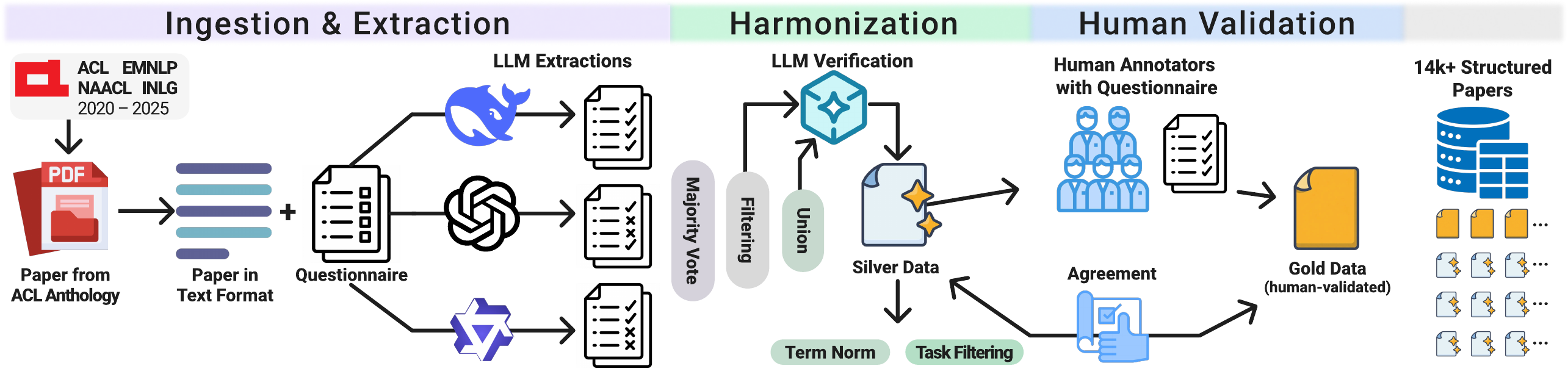}
    }
    \caption{
        Our paper annotation pipeline (\S \ref{sec:annotation}), including converting PDF to text (\S \ref{sec:extraction}), extraction of metadata based on four NLG evaluation questions (Table~\ref{tab:llm-schema}), harmonization based on majority vote, filtering, and merging (\S \ref{sec:merging}), and human validation (\S \ref{sec:human_annotation}), yielding 14k structured papers including a subset of 110 papers with gold annotations.
    }
    \label{fig:methodology}
\end{figure*}

\section{Paper Annotation}
\label{sec:annotation}

We include main conference papers from the ACL Anthology from 2020 to 2025, focusing on four conferences: ACL, EMNLP, NAACL, and INLG.
In total, 14,171 papers were collected (full statistics in Table~\ref{tab:conference_stats} of Appendix~\ref{append:results}). 
Next, we describe our paper annotation process as shown in Figure~\ref{fig:methodology}.

\subsection{LLM Annotation}
\label{sec:extraction}
We extract the full text of each paper from PDF format using GROBID\footnote{\url{https://github.com/kermitt2/grobid}}. We then feed the text (excluding title and abstract, as the full text should contain all information we need) into a model with our designed prompt.
For each paper, we prompted an LLM acting as an ``expert NLP researcher with deep experience in Natural Language Generation'' to read the text and return a structured JSON object answering four binary questions about NLG-related information: (Q1) Does the paper address NLG tasks? (Q2) Does the paper use automatic metrics to evaluate the generated outputs? (Q3) Does the paper use large language models (LLMs) as judges (i.e., after generation, LLMs are used to assess the outputs)? (Q4) Does the paper conduct human evaluations of the generated~outputs?

If the \texttt{``answer''} to a question is \texttt{``Yes''}, we ask the LLM to extract further metadata, e.g., a verbatim excerpt as evidence, lists of datasets, languages, models, and criteria. If the \texttt{``answer''} is \texttt{``No''}, all other fields in that section are set to empty. The prompt and full metadata list are provided in Appendix~\ref{appendix:prompt1} and Table~\ref{tab:llm-schema}.

We categorize and define the evaluation methods by 
(1) automatic metrics: metrics used to measure generated text quality by providing a score; 
(2) LaaJ: evaluating generated text with LLM prompting, producing textual evaluation and a rating score; and 
(3) Human evaluation: evaluation produced by humans following a given guideline. 

\paragraph{Model Selection}

We performed the extraction using three different open-source models: DeepSeek-R1 \cite{deepseekai2025deepseekr1incentivizingreasoningcapability}, GPT-OSS-120B \cite{openai2025gptoss120bgptoss20bmodel} and Qwen3-235B-A22B-Instruct \cite{yang2025qwen3technicalreport}. 

\subsection{Harmonization of LLM-based Extraction}
\label{sec:merging}
We aggregate the results produced by the three LLMs to create a unified final output. We use majority voting on the four binary questions, and the union of three metadata lists. We filter out papers that have \textit{no} to Q1, resulting in 8,665 initial NLG papers (61\%). For these papers, we further employed another LLM (DeepSeek V3.1 Terminus) to verify the extraction with another prompt. The prompt is designed to verify each extraction against the full paper text to validate the four binary questions, and perform four actions: (1) \textbf{Normalize} metadata to use canonical forms (e.g., all variants of BLEU (bleu, Bleu, BLEU-4, etc.)); (2) \textbf{Correct} any incorrect items correct; (3) \textbf{Add} any missing important items; and (4) \textbf{Remove} any irrelevant or incorrect items from the list for each field listed in Table~\ref{tab:llm-schema} (see full prompt in Appendix~\ref{appendix:prompt2}).

\paragraph{Comparison of LLM annotations}
To compare the difference among LLM annotations, we compute the agreement with Krippendorff's $\alpha$ (pairwise agreement between the LLM-harmonized one and the others, shown in Table~\ref{tab:pairwise-agreement} in Appendix). The overall agreements are high for all four answers (Q1: 0.710, Q2: 0.688, Q3: 0.805, and Q4: 0.812).

\subsection{Extraction of LaaJ-Human Validation}
\label{sec:prompt3}

To determine how many papers validated LaaJ against human evaluation, we analyzed 433 NLG papers (12.3\% of 3,334 filtered NLG papers detailed in \S\ref{para:nlg_filtering}) that employed both methods. Because our initial pipeline did not extract comparison metadata, we designed a supplementary prompt using DeepSeek V3.1 Terminus (Appendix~\ref{appendix:prompt3}). This prompt extracts: (1) a binary indicator of whether a comparison exists; (2) the comparison method (e.g., correlation, agreement); (3) the statistical metrics used (e.g., Pearson, Spearman, accuracy); and (4) the quantitative results (e.g., correlation scores, significance). Crucially, we required all extracted results to be mapped to specific evaluation criteria.

\subsection{Post-processing Extractions}

\paragraph{Term Normalization} 
To group term variants referring the same value, we perform term normalization. For tasks, datasets, models, languages, and automatic metrics, we apply two normalization steps. (1) Spelling unification and (2) Term merging with fuzzy-matches to merge near-duplicates (e.g., \textit{Dialog Generation} $\rightarrow$ \textit{Dialogue Generation}). 
For detailed term normalizations and examples, please check Appendix~\ref{append:term_normalization} and \ref{append: term_norm_examples}.

\paragraph{Criteria Mapping via QCET}
Because fuzzy matching often over-merges distinct criteria (e.g., \textit{factual correctness} vs.\ \textit{factual consistency}), we instead classified each raw criterion into the QCET taxonomy~\citep{belz-2024-qcet} (a 111-leaf lattice structured by frame-of-reference $\times$ type $\times$ aspect) using a two-step LLM pipeline. In Step 1, the LLM mapped each criterion to existing QCET leaves, assigning a match confidence of ``strong'', ``partial'', or ``none''. For criteria lacking a strong match, we clustered their embeddings using HDBSCAN to identify gaps in the taxonomy. This revealed four new criteria clusters, to which we added two more based on high overall frequency. We mapped these six new leaves to compatible parent branches (Table~\ref{tab:qcet_extensions}), expanding the taxonomy to 117 leaves. We also introduced two auxiliary categories: \textit{Overall Quality} (for holistic judgments) and \textit{Other / Unclassifiable} (which is dropped from analysis). In Step 2, the LLM re-classified all ``partial'' and ``none'' criteria against this extended taxonomy, using their Step 1 labels and embedding clusters as additional signals. 
To validate this mapping, two annotators evaluated a stratified random sample of 155 criteria on a 1--5 Likert scale (1=wrong, 5=perfect). The average scores from annotators were 4.45 and 4.81, indicating an acceptable mapping.
More implementation details are in Appendix~\ref{append:term_normalization}.

\paragraph{NLG Task Filtering}
\label{para:nlg_filtering}

Our human annotations (\S \ref{sec:human_annotation}) reveal that most disagreements regarding Q1 (NLG vs. non-NLG) come from papers involving either classification tasks or generation tasks that produce non-natural-language outputs. 
We narrow our scope of NLG as tasks that produce open-ended natural language outputs evaluated on linguistic and semantic criteria (e.g., fluency, coherence, and relevance), and deliberately exclude classification tasks and generation tasks where outputs are artifacts verifiable by execution or exact correctness (detailed reasons for these exclusions are provided in Table~\ref{tab:excluded_tasks}).
After normalizing and ranking the tasks by frequency, we systematically exclude 15 tasks (see Appendix~\ref{append:task_filtering} for the full list) and retain the top 30 tasks for our analysis (3,334 papers), shown in Figure~\ref{fig:task_top}. 

\section{Human Validation of Extractions}
\label{sec:human_annotation}

\begin{table}[h]
    \centering
    \begin{adjustbox}{width=\columnwidth}
    \begin{tabular}{lcccc}
        \toprule
        Pair & Q1 & Q2 & Q3 & Q4 \\
        \midrule
        Human R1 vs. Human R2        & 81.82 & 80.00 & 89.09 & 91.82 \\ \midrule
        Human vs. DeepSeek-R1     & 74.55 & 73.64 & 95.45 & \textbf{92.73} \\
        Human vs. GPT-OSS-120B    & 75.45 & \textbf{75.45} & 90.91 & 90.91 \\
        Human vs. Qwen3-235B      & 75.45 & 70.00 & 93.64 & 85.45 \\
        Human vs. Majority Voting & 74.55 & 71.82 & 93.64 & 88.18 \\ 
        Human vs. LLM-harmonized      & \textbf{77.27} & 74.55 & \textbf{95.45} & 90.00 \\ 
    \bottomrule
    \end{tabular}
    \end{adjustbox}
    \caption{Pairwise agreement (\%) between human annotations and LLMs.}
    \label{tab:human_vs_llm}
\end{table}

\subsection{Validation of LLM extractions}

\paragraph{Human annotations}
To assess the quality of the LLM extractions, we manually annotated 110 randomly selected papers that were identified as NLG research by a majority LLM vote on Q1. Our annotation guidelines were modeled after the LLM prompts, but included additional details and examples (Appendix~\ref{append:human}). The process was conducted by 11 NLP researchers (3 Master's students, 2 PhD students, and 6 Postdocs). To ensure reliability, each paper received two independent annotations, with any disagreements resolved by a third annotator.
With the human-annotated ground truth, we compare LLM-extractions against it.

\paragraph{Binary-answer agreement}
When evaluated against the human ground truth on the four binary questions (Table~\ref{tab:human_vs_llm}), Q3 and Q4 demonstrate higher agreement than Q1 and Q2. This mirrors the inter-LLM agreement (Table~\ref{tab:pairwise-agreement}) and suggests that the adoption of LaaJ and human evaluation is easier to identify than the use of NLG tasks and automatic metrics. Notably, the harmonized LLM output (\S\ref{sec:merging}) achieves the highest overall agreement with human, validating our harmonization step.

The lower agreements on Q1 and Q2 are mainly due to lower precision (Q1-Q4: $0.78$, $0.72$,  $0.84$, $0.75$), as the recall is near-perfect on all four questions ($0.98$, $0.96$, $1.0$, $1.0$). This means extraction errors are almost exclusively false positives, consistent with our high-recall design goal of not missing NLG papers. Moreover, these false positives are coupled: most Q2 and Q4 errors occur on papers the annotators classified non-NLG (which entails ``no'' downstream); among papers where human and LLM agree on Q1, agreement rises to $92.5\%$ (Q2) and $96.2\%$ (Q4), and the NLG task filtering removes many of the remaining Q1 false positives.

\paragraph{Metadata-level extraction quality}
To inspect extraction beyond the binary answer level, we sampled 60 of the 110 papers and asked annotators to score how closely each LLM-extracted metadata matches their own annotation, on a 1--3 Likert scale (1=mismatch, 2=partial match, 3=full match). All 11 fields exceed an average score of 2.5, ranging from 2.52 (NLG models, output formats) to 2.91 (LaaJ models). LaaJ models, languages, and LaaJ criteria score highest; NLG models and output formats lowest (Appendix Table~\ref{tab:likert_extraction}).

\subsection{Validation of non-NLG paper filtering}
To test whether LLMs correctly filter out non-NLG papers, we randomly sampled 120 papers from the filtered-out ``non-NLG'' set; 6 NLP-researcher annotators each labelled 40 papers, and each paper received two annotations. Both annotators agreed with the rejection for 117 of the 120 papers (97.5\%), with only three disagreements, confirming that LLMs correctly filter out non-NLG papers in almost all cases.

\subsection{LaaJ-human validation extraction quality}
To evaluate the supplementary LLM extractions for LaaJ-human validation, nine NLP researchers annotated a sample of 90 papers. For binary classification (presence of validation), human annotations achieved 88.9\% agreement with the LLM. However, when extracting fine-grained criterion comparisons, humans identified an average of 13.86 distinct data points per paper compared to the LLM's 4.31. Our manual inspection indicates that under-extraction is caused by PDF-to-text conversion rather than the LLM. Tabular content is largely lost during parsing, while text mentions of aggregate scores survive. Therefore, we use LLM extraction for presence of validation and rely on human annotation for the per-criterion alignment analysis in \S\ref{sec:laaj_human_subsec}. More evaluation details are in Appendix~\ref{append:human}.

\section{Results and Analysis}
\label{sec:results}

To quantitatively analyze the extracted results, we use two co-occurrence measures defined over pairs of terms (a term being a task, a metric, or a LaaJ/human evaluation criterion). \textbf{Frequency} $P(B|A)$ is the proportion of papers containing term $A$ that also contain term $B$; it captures how widely $B$ is used within the subset of papers that use $A$. The \textbf{Likelihood Ratio} (LR) $LR(A\to B) = P(B | A) / P(B | \neg A)$ measures how much more likely $B$ is to appear in papers using $A$ than in papers not using $A$, capturing the distinctiveness of the co-occurrence. For example, in our data $81.5\%$ of MT papers report BLEU ($P(\text{BLEU} | \text{MT}) = 0.815$), and BLEU is $2.6\times$ more likely in MT papers than in non-MT papers ($LR(\text{MT} \to \text{BLEU}) = 2.63$). Despite the shared name, our LR is the relative-risk ratio from epidemiology, not the statistical likelihood-ratio test; full formulas for the three LR variants used in this paper appear in Appendix~\ref{append: lr_formula}.

\paragraph{Statistical significance testing}
Because LR can be inflated for rare co-occurrences, we pair it with the Dunning $G^2$ log-likelihood ratio test~\citep{dunning1993}.
We then apply Benjamini--Hochberg FDR correction~\citep{benjamini1995} across all tested pairs within each analysis family and retain only pairs with corrected $p \leq 0.05$. 

\subsection{Temporal Trend Analysis}

\begin{figure}[ht]
    \centering
    \includegraphics[width=\columnwidth]{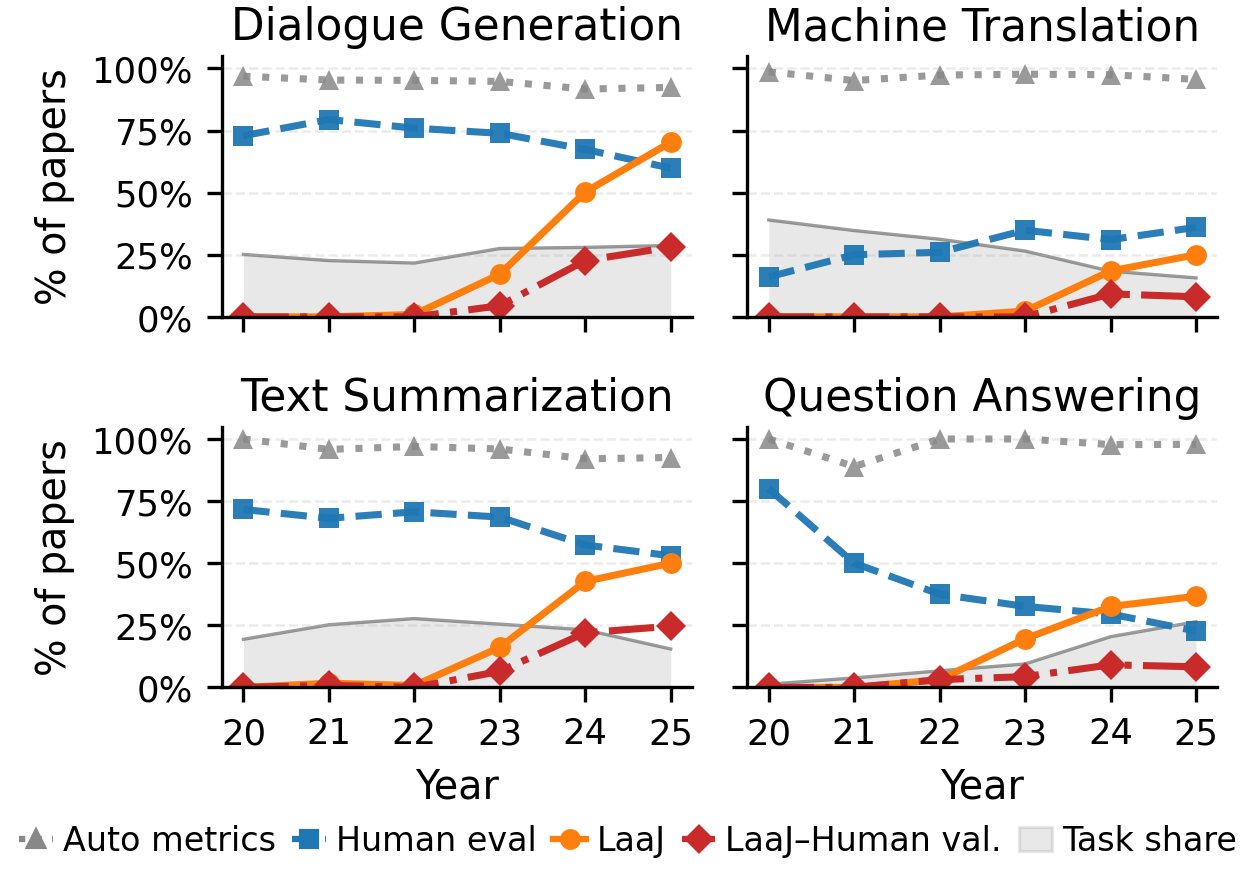}
    \caption{Per-task evaluation method adoption (\%) of top-four NLG tasks per year. The \textit{filled area} shows that task's share of all NLG papers each year.}
    \label{fig:task_four}
    \vspace*{-0.5em}
\end{figure}

Figure~\ref{fig:task_four} presents the evolution of evaluation practices across the four most prominent NLG tasks, which account for 78.1\% of the papers: Dialogue Generation (DG; 26.1\%), Machine Translation (MT; 25.4\%), Text Summarization (TS; 22.0\%), and Question Answering (QA; 13.9\%). We observe three key trends. First, automatic metrics remain universally applied (near 100\%) across all tasks. Second, traditional tasks like MT rely almost entirely on these automatic metrics, with only 26\% conducting human evaluation and negligible LaaJ adoption until 2025. Third, in tasks like DG, TS, and QA, we observe a potential substitution effect: as LaaJ adoption has rapidly accelerated since 2023, human evaluation has either dipped (DG, TS) or plummeted entirely (QA). Importantly, across all tasks, the validation of LaaJ against human evaluation remains relatively low compared to its usage (\S\ref{sec:laaj_human_subsec}).

\begin{figure*}[h]
    \centering
    \includegraphics[width=\linewidth]{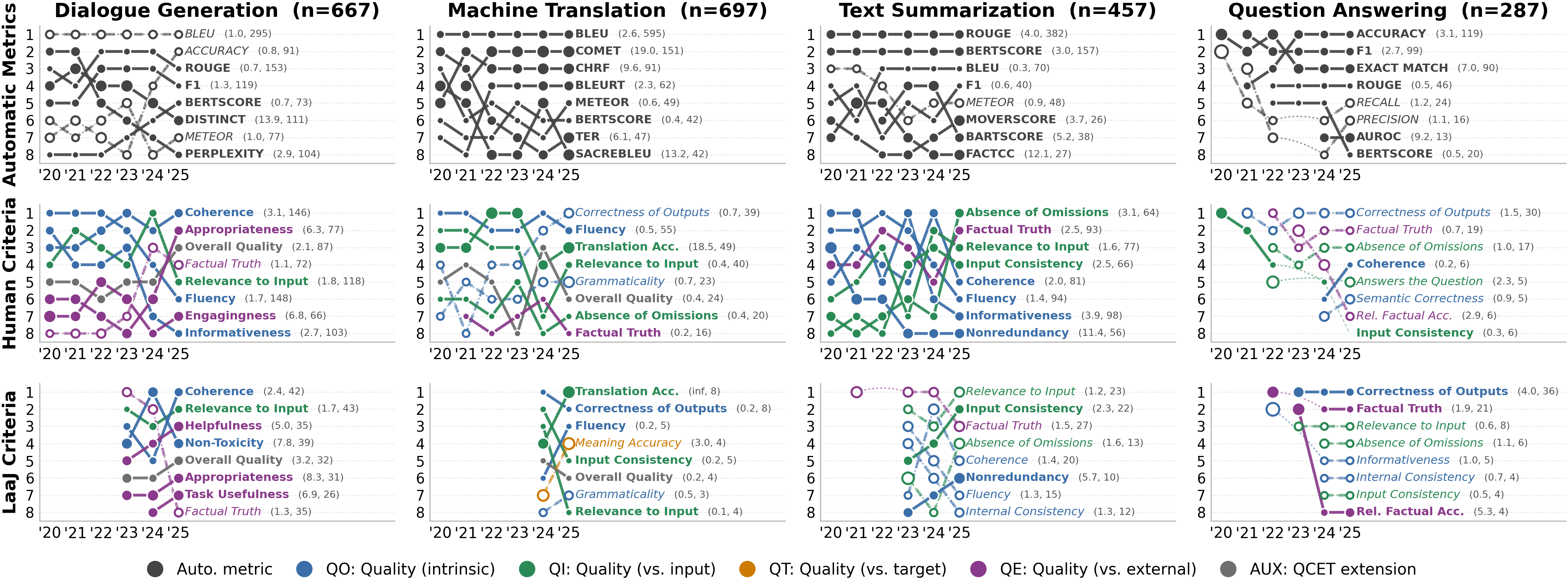}
    \caption{Evaluation trends for the four most frequent NLG tasks (columns) under three evaluation paradigms (rows). Each panel is a separate bump chart: the $y$-axis gives the annual frequency rank (1 = most used) of the eight most common metrics for that task, from 2020 to 2025 on the $x$-axis. Marker size scales with the metric's likelihood ratio (LR) for the task in that year. Solid lines with filled markers mark a significant overall association with the task; dashed lines with hollow markers mark metrics that are frequent but have a low LR. Endpoint labels give the metric name and \texttt{(overall LR, $k$)}, where $k$ is the number of papers. COMET's \texttt{(19.0, 151)} in the MT/Automatic Metrics panel thus means it appears in 151 MT papers and is 19 times more likely to appear in MT than elsewhere; it enters at rank 4 in 2021 and holds rank 2 from 2022 onward, behind BLEU. Dotted arcs bridge years in which a metric drops out of the top eight; dashed connectors attach labels to metrics that are outside the top eight in 2025. A missing line in a given year means the metric was not used. Color denotes the QCET L1 dimension: QO (intrinsic to output), QI (relative to input), QT (relative to target reference), QE (relative to external standard), and AUX (QCET extensions).}
    \label{fig:task_year}
    \vspace*{-1em}
\end{figure*}

We further examine each of the top four tasks and discuss how the evaluation methods differ. We restrict the analysis to single-task papers to find task-specific patterns, and visualize the top 8 most frequent metrics and criteria for that year. Figure~\ref{fig:task_year} visualizes the evolution of evaluation methods across the four major tasks. 

\vspace*{-0.5em}
\paragraph{Dialogue Generation} 
As Dialogue Generation (DG) shifts from simple request fulfillment toward complex, open-ended interactions, its evaluation landscape has correspondingly diversified (Figure~\ref{fig:task_dashboard}). Among automatic metrics, \textit{BLEU} remains the most frequent, though it exhibits a low task association (LR). Conversely, \textit{Distinct}, which measures diversity, shows the highest LR with DG. Traditional performance metrics like \textit{Accuracy} and \textit{F1} also exhibit low LRs, likely diluted by their increasing use in other tasks. For human evaluation, \textit{Engagingness} and \textit{Appropriateness} are the most strongly associated criteria, even though \textit{Coherence} is used more frequently overall. Notably, the reliance on \textit{Fluency} and \textit{Informativeness} has declined sharply, likely reflecting the baseline performance gains of modern LLMs. When comparing paradigms, both LaaJ and human evaluation emphasize \textit{Appropriateness}. However, LaaJ uniquely prioritizes alignment-focused criteria such as \textit{Helpfulness} and \textit{Non-Toxicity}. This indicates that LaaJ is introducing novel safety dimensions, revealing a potential gap in human evaluation practices.

\vspace*{-0.5em}
\paragraph{Machine Translation}
MT primarily uses \textit{BLEU} (87.4\%) for evaluation (while rather low association (since it is also heavily used in other tasks), with \textit{COMET} also established as a major metric. The latter also has the highest association (LR=19) with MT since 2023. 
In terms of evaluation criteria, \textit{Correctness of Outputs} and \textit{Fluency} are among the most frequent ones, despite their low LR. \textit{Translation Accuracy} is ranked highest in association (the criterion is unique to the MT task). 
The dominance of these n-gram and reference-based metrics illustrates deep-seated inertia (formally defined in \S \ref{sec:challenges}). 
Despite the availability of LLMs, the LaaJ Criteria row is sparse, without any activity until 2024.
Overall, the evaluation paradigm in MT has not shifted as much as the other tasks.

\vspace*{-0.5em}
\paragraph{Text Summarization}
In TS, \textit{ROUGE} and \textit{BERTScore} dominate automatic evaluation, both showing high task associations (LRs). \textit{FactCC} has an increased 2021--2022 frequency for evaluating hallucinations and retains the highest LR despite a recent drop. Notably, \textit{BLEU} is frequently used but lacks significant task association. In human evaluation, \textit{Nonredundancy} exhibits the highest LR despite low usage, while \textit{Absence of Omissions} and \textit{Factual Truth} are increasingly prioritized over declining criteria like \textit{Fluency} and \textit{Informativeness}. Lastly, most LaaJ criteria currently show no significant statistical association with TS.

\vspace*{-0.5em}
\paragraph{Question Answering}
Because QA typically generates relatively short responses, its evaluation paradigm diverges substantially from the other three tasks. It often resembles a classification problem, relying heavily on exact-match comparisons against a ground truth. Consequently, QA exhibits the least diverse set of evaluation metrics and criteria. Most criteria have no significant LR with QA, suggesting a lack of task-specific evaluation designs. The predominant criterion is \textit{Correctness of Outputs}, particularly when employing LaaJ. This trend is concerning: although QA has experienced the most rapid growth, its evaluation remains heavily reliant on simplistic string matching, misaligned with the complexities of open-ended~QA.

\subsection{Metric-criteria Association by Evaluation Methods}
\label{subsection: metric-evaluator}

\begin{figure*}[t]
    \centering
    \includegraphics[width=\textwidth]{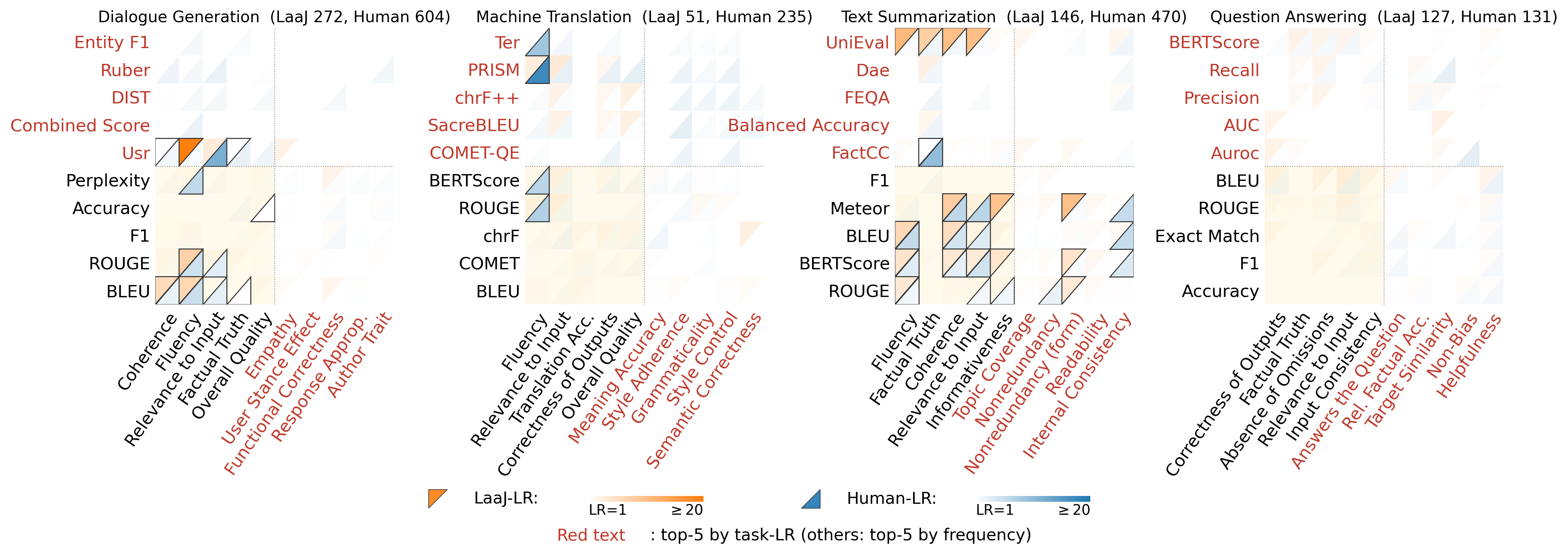}
    \caption{Within-task co-occurrence LRs for metric--criterion pairs, one panel per task. Rows list metrics and columns list criteria; each axis contains the five most frequent items (black labels) and the five with the highest task-LR (red labels), so the quadrant where frequent metrics meet frequent criteria (yellow background) is distinct from the rest. Each cell is split along the diagonal: the upper-left orange triangle covers the task's LaaJ papers and the lower-right blue triangle its human-evaluation papers. Colour intensity encodes the co-occurrence LR, and an outline marks a significant pair. In Dialogue Generation, BLEU with Fluency is outlined and shaded on both halves (133 human-evaluation papers, $\mathrm{LR}=2.02$; 19 LaaJ papers, $\mathrm{LR}=2.59$), indicating that the pairing is associated with the task under both paradigms, whereas a cell shaded on one side only reflects a paradigm-specific pairing. Across tasks, papers pair metrics and criteria along frequently used terms rather than task-distinctive ones.}
    \label{fig:metric_methods}
    \vspace*{-0.5em}
\end{figure*}

Although task-specific LRs identify the metrics and criteria most associated with each NLG task (Figure~\ref{fig:task_year}), this term-level association does not imply paper-level pairing. A highly associated metric and criterion may both be common in a task's corpus without ever co-occurring in the same paper. 
To study their co-occurrence and association level, we compute the within-task pair association LR for every (metric, criterion) pair, as most metrics do not explicitly mention the criterion it measures\footnote{We manually annotated 50 randomly sampled papers containing automatic metrics (183 metric mentions, two
annotators, $\kappa=0.49$) to confirm this. Only $21$--$36\%$ of metric
mentions explicitly state which criterion the metric measures, and $6\%$
of papers state a mapping for every metric
(Appendix~\ref{append:mapping}).}. We calculate separately for the LaaJ and human evaluation subsets of the task's papers (Figure~\ref{fig:metric_methods}).

Significant metric-criterion associations concentrate almost exclusively in the frequency $\times$ frequency quadrant. Across DG and TS, \textit{BLEU}, \textit{ROUGE}, \textit{BERTScore}, and~\textit{METEOR} pair significantly with \textit{Coherence}, \textit{Fluency}, and \textit{Relevance to Input}, indicating a default evaluation set shared across tasks. The distinctiveness $\times$ distinctiveness quadrant contains no significant cells: task-distinctive metrics and criteria are well attested individually but rarely co-reported. A few distinctive metrics do reach high LR against frequent criteria. In TS, \textit{FactCC} pairs near-exclusively with \textit{Factual Truth}, consistent with its design as a factuality metric. The orange-versus-blue split of these cells is itself informative: \textit{UniEval}'s associations reach significance only among LaaJ papers, while MT's four significant cells are all human-side and all involve \textit{Fluency}; indicating a divergence between LaaJ and human evaluation paradigms.
MT and QA show much weaker structure: no metric in MT reaches significance against the task-distinctive criteria (\textit{Translation Accuracy}, \textit{Meaning Accuracy}, \textit{Grammaticality}); QA yields no significant pairings under either method.

These results indicate that NLG papers choose their metric and criterion largely independently, each defaulting to frequent measures rather than pairing task-specific ones. This decoupling raises a general construct-validity concern: the chosen metrics rarely reflect a deliberate methodological match to the reported criteria.

\begin{figure*}[!h]
    \centering
    \includegraphics[width=\textwidth]{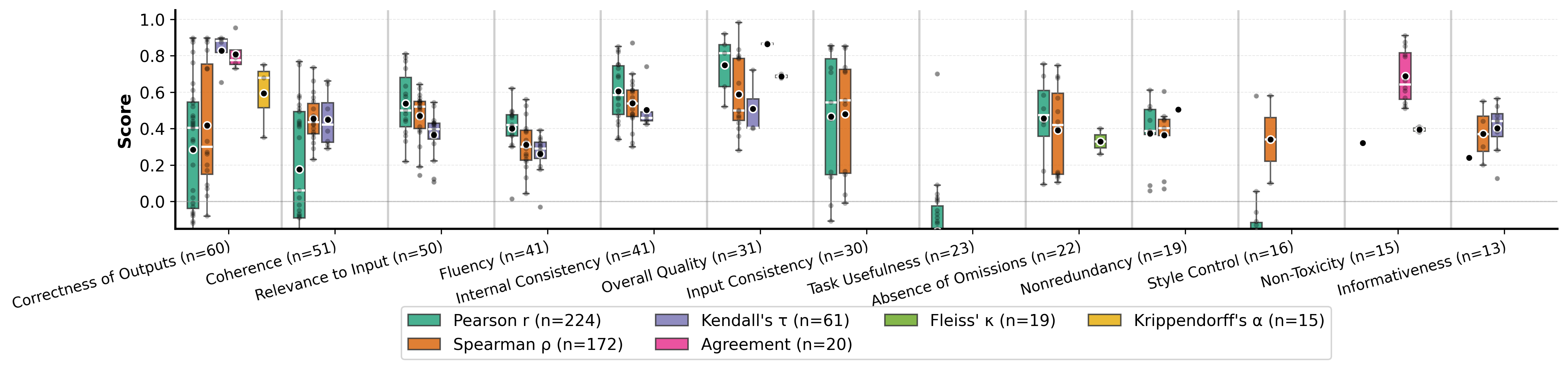}
    \caption{Distribution of LaaJ vs. human evaluation scores by metric and criterion. Individual observations are shown as gray dots. $n$ denotes the number of studies included. Values are based on human annotated 70 papers.}
    \label{fig:laaj_human}
    \vspace*{-1em}
\end{figure*}

\subsection{LaaJ with Human Validation}
\label{sec:laaj_human_subsec} 

Using the extraction methodology from \S\ref{sec:prompt3}, we extracted validation results from 433 papers employing both LaaJ and human evaluation. Only 254 of these (58.7\%) explicitly validate LaaJ against human judgments, representing $37.4\%$ of the 679 NLG papers using LaaJ, under 8\% of total 3,334 NLG papers. Among the remaining 179 papers without an explicit comparison, $76.3\%$ still measure at least one shared criterion across LaaJ and human, suggesting comparison was generally feasible but unreported, rather than the two methods deliberately covering disjoint aspects. 
Figure~\ref{fig:laaj_human} visualizes per-criterion alignment using the 70 human-annotated subset of papers (\S\ref{sec:human_annotation}), restricted to QCET-mapped criteria with more than $10$ comparison values.
As shown in Figure~\ref{fig:laaj_human}, evaluating our human-annotated subset reveals a disparity in LaaJ reliability. While automated judges achieve moderately high median correlations for aggregate \textit{Overall Quality}, their alignment with human annotators degrades significantly on nuanced, fine-grained criteria. Dimensions such as \textit{Fluency} and \textit{Coherence} exhibit median correlations near or below 0.4, accompanied by large variance, demonstrating that LaaJ struggles to reliably capture targeted linguistic constructs. Furthermore, as often cautioned in the NLG community, evaluating surface-level quality is insufficient for determining real-world utility; our data confirms that LaaJs fail to align with humans on \textit{task-specific usefulness}.

\subsection{Analysis on the Decline of Human Evaluation}

To investigate the decline of human evaluation, we disentangle which factors explain for the presence of human evaluation with a regression analysis. We model presence of human evaluation as a dependent variable; year, task (top-four tasks), venue and presence of LaaJ as the independent variables. Specifically, we report LR with 95\% confidence intervals (a CI interval excluding 1 indicates significance) with modified Poisson regression with robust standard errors, as the presence of human evaluation is binary. With a single variable, the regression reduces exactly to LR; with the several variables listed, it yields each variable’s LR adjusted for the others. 

\begin{table}[!t]
    \centering
    \small
    \begin{tabular}{ll}
        \toprule
        \textbf{Variable (reference)} & \textbf{LR [95\% CI]} \\
        \midrule
        LaaJ use & \textbf{1.28} [1.16, 1.41] \\
        Year 2021 (vs. 2020) & 1.11 [0.98, 1.27] \\
        Year 2022 & 1.08 [0.95, 1.23] \\
        Year 2023 & 1.12 [0.99, 1.27] \\
        Year 2024 & 0.97 [0.84, 1.11] \\
        Year 2025 & \textbf{0.86} [0.74, 1.00] \\
        Machine Translation & \textbf{0.49} [0.42, 0.56] \\
        Text Summarization & \textbf{1.15} [1.02, 1.30] \\
        Dialogue Generation & \textbf{1.16} [1.02, 1.31] \\
        Question Answering & \textbf{0.51} [0.43, 0.61] \\
        Venue EMNLP (vs. ACL) & 0.97 [0.90, 1.05] \\
        Venue INLG & 1.08 [0.91, 1.28] \\
        Venue NAACL & 1.04 [0.94, 1.15] \\
        \bottomrule
    \end{tabular}
    \caption{Regression result. Bold: 95\% CI excludes 1, significant. Year and venue are exclusive categories, compared to a reference (2020 / ACL); task indicators are not exclusive (a paper can address several tasks), so each task LR compares papers with vs. without that task}
    \label{tab:lr_variables}
    \vspace*{-1.15em}
\end{table}

From the regression results, we can explain how each variable explains the presence of human evaluation: (1) ``Task'' strongly affects the presence of human evaluation, MT and QA papers are half as likely to conduct it compared to TS and DG, consistent with our findings; (2) ``Venue'' has no significant impact on human evaluation trends; (3) ``Year'' has a significant impact in 2025, consistent with our claim that human evaluation declines in 2025; (4) ``LaaJ'' results show that papers using LaaJ are more likely to report human evaluation, thus increase of LaaJ usage does not mean its substitution of human evaluation. 

\section{Challenges and Guidelines}
\label{sec:challenges}

\vspace*{-0.5em}
\paragraph{Metric Inertia}
A persistent challenge in NLG evaluation is the reluctance to discontinue lexical-overlap based metrics, a pattern we term metric inertia. Part of this inertia is driven by the structure of legacy datasets, which typically provide a single ``gold standard'' reference. Researchers default to legacy metrics because they offer a fast, deterministic way to calculate overlap against this ground truth. However, modern LLMs excel at stylistic variation, paraphrasing, and generating valid alternative answers that do not share the exact vocabulary of a single human annotator \cite{liu-etal-2021-language}. By continuing to anchor evaluations to lexical overlap, NLG field risks optimizing for surface-level reference mimicry rather than actual semantic quality. Even when a ground-truth reference is available, researchers should prioritize reference-based semantic metrics (e.g., BERTScore, BLEURT) that reward meaning over rigid vocabulary matching.

\vspace*{-0.5em}
\paragraph{Metric-Criteria Mapping}
Beyond temporal inertia, automatic metrics are applied across both tasks and criteria. Cross-task, legacy lexical metrics like BLEU and ROUGE are used across all four focus tasks with weak LR (Figure~\ref{fig:task_year}); within a task, they pair with multiple frequent quality criteria (Figure~\ref{fig:metric_methods}). Researchers therefore select metrics based on popularity or tradition rather than the construct under evaluation, creating conceptual entanglement and confirming the lack of standardization noted by \citet{belz-etal-2025-standard}. The widespread reporting of established metrics as generic proxies for ``quality'' limits their interpretability \cite{sai-2022-evaluation-metrics-nlg-systems}; without explicitly binding a metric to a specific linguistic or semantic property, it is difficult to determine which distinct aspect of generation is driving an observed improvement.

\vspace*{-0.5em}
\paragraph{Validation Gap}
The accelerating adoption of LLM-as-a-Judge (LaaJ) has outpaced its empirical validation. Our data indicates that while LaaJ usage has now surpassed human evaluation (Figure~\ref{fig:eval_method_adoption}), direct comparative validation on the same data remains infrequent. More notably, the existing validation evidence skews heavily toward aggregate measures like \textit{Overall Quality} and \textit{Correctness of Outputs}, while demonstrating high variance and weaker alignment on fine-grained criteria. This validation gap introduces a methodological vulnerability: by increasingly relying on automated evaluators without construct-specific human baselines, we risk standardizing benchmarks that may silently diverge from human perception on nuanced dimensions like \textit{Task usefulness} (Figure~\ref{fig:laaj_human}).

\vspace*{-0.5em}
\paragraph{NLG Evaluation Checklist}
To help researchers translate these challenges into practice, we propose a compact Evaluation Design Checklist and task-specific recommendations for DG, MT, TS, and QA (Appendix~\ref{append:recommendations}).
\begin{guidelinebox}
\footnotesize  
\noindent\textit{1. Metric Selection}
\begin{itemize}[nosep, leftmargin=*, label={$\square$}]
    \item \textbf{Are you evaluating open-ended generation?} If yes, acknowledge that optimizing for exact n-gram overlap with a single reference penalizes valid paraphrases. Shift legacy metrics to secondary baselines.
    \item \textbf{Have you included a semantic metric?} Ensure legacy metrics (BLEU, ROUGE) are supplemented, or replaced, by semantic metrics (e.g., BERTScore, COMET, BLEURT) that reward meaning over lexical matching.
\end{itemize}

\vspace{0.3em}
\noindent\textit{2. Construct Validity}
\begin{itemize}[nosep, leftmargin=*, label={$\square$}]
    \item \textbf{Is the target construct explicitly named?} Verify that every reported metric is explicitly tied to a specific quality criterion (e.g., ``We use QAFactEval to measure factuality,'' not just ``to measure quality'').
    \item \textbf{Is the metric mathematically suited for the construct?} Ensure you are not using lexical overlap (e.g., ROUGE) to substantiate claims about high-level semantics (e.g., logical coherence, factual consistency).
\end{itemize}

\vspace{0.3em}
\noindent\textit{3. LaaJ Deployment}
\begin{itemize}[nosep, leftmargin=*, label={$\square$}]
    \item \textbf{Have you validated your LaaJ against human judgments?} Where a public meta-evaluation benchmark covers your target criterion, report your judge's correlation against it. Otherwise, annotate a subset per criterion and report appropriate correlation or agreement metrics.
    \item \textbf{Are human experts retained for sensitive or poorly correlated criteria?} Acknowledge that LaaJ reliability often collapses on strict linguistic nuances (e.g., \textit{task usefulness}). For these criteria, rely on expert human evaluation as your primary signal.
    
\end{itemize}
\end{guidelinebox}

\section{Conclusion}
We present a large-scale quantitative analysis of 3,334 NLG papers (2020--2025) across four NLP conferences, utilizing a human-validated, multi-LLM extraction pipeline. Our analysis indicates that NLG evaluation methodologies are failing to keep pace with the capabilities of modern generation models. We highlight three recurrent issues: the persistence of metric inertia, metric-criteria mapping problem, and a severe validation gap in LaaJ evaluation. Crucially, our human-annotated subset demonstrates that while LaaJ approximates aggregate quality, its alignment with humans degrades severely on fine-grained criteria. To ensure evaluation protocols mature alongside NLG capabilities, we contribute an Evaluation Design Checklist, urging researchers to explicitly map metrics and rigorously validate LaaJ on fine-grained dimensions.

\section*{Limitations}
We list a few limitations that should further improve the paper in future research:

\textbf{Multi-task papers}: some papers deal with multiple tasks; however, our annotation guideline did not require users/LLMs to extract metadata per task (models, language, and evaluation metrics), thus, there can be some deviations in reported numbers. 

\textbf{Definition boundary of metrics and LaaJ}: the definition of automatic metrics and LaaJ is not clearly separable. Some learned metrics that produce a score may have been considered as LaaJ by LLM-extraction; while performing annotation, humans do not always reach an agreement. 

\textbf{Task granularity}: tasks such as Dialogue Generation are still quite general; under them, there can be many subtasks in which very different metrics and criteria are used. Our association cannot represent these more specific tasks.

\textbf{Extracted metadata evaluation}: Our human evaluation of LLM-extraction on the metadata-level assigns a 1-3 likert scale (1=mismatch, 3=match); however, more fine-grained criteria could improve the evaluation, for example, by asking the type of errors (missing item, hallucinated item or paraphrased item). We leave this to future work for improving LLM information extraction.

\textbf{Criteria mapping}: when mapping the raw extracted criteria to the QCET taxonomy, the input to LLMs is only the criterion name with no context information; further information about the context can be extracted from papers (although some papers might not have such context, such as definition of their criteria) may help with better mapping.

\section*{Ethical Statement}
The human annotators are NLP researchers who are either co-authors of this paper or students who are hired by their host institutions. 
AI (Claude) was used for coding (computing LR and agreements) and figure polishing, and in writing for grammar and sentence coherence improvement. All AI-assisted content were reviewed and verified by the authors. 

\section*{Acknowledgments}
We thank the anonymous reviewers at ACL Rolling Review for their valuable feedback. We also thank the additional human annotators (Awsam Agbaria, Alexandra Tsiakalou, Jiaao Li, Neda Foroutan) for their valuable time and effort, and our colleague Shushen Manakhimova for her constructive comments on the manuscript. This research is funded by the Berlin Institute for the Foundations of Learning and Data (BIFOLD) and the Federal Ministry of Education and Research (BMFTR) in the scope of the research project VeraXtract (ref. 16IS24066) and  GenKI4Media (ref. 01MK25004E).

\bibliography{custom}

\appendix

\section{License of Artifacts}
Our data are based on open-source research papers from ACL Anthology, which are licensed to the general public under a liberal usage policy that allows unlimited reproduction, distribution and hosting of materials on any other website or medium, for non-commercial purposes. 
LLMs we used are all open-source models, which all permit the usage for research purposes.

\section{Experimental Details}

\subsection{Model Setup}
Due to large model sizes, we used DeepSeek official API for DeepSeek-R1 and Novita API~\footnote{\url{https://novita.ai/}} for the other LLMs inferences. We set temperature as 1 for all models. Total cost is around 50 dollars. 

\subsection{Term Normalization}
\label{append:term_normalization}

\paragraph{Task}
To normalize the task names, we perform three steps: (1) removing punctuation and separators like /, \& and -; (2) replacing acronyms with their standard forms (e.g., QA $\rightarrow$ Question Answering); and (3) applying fuzzy matching to merge near-duplicates (e.g., Dialog Generation $\rightarrow$ Dialogue Generation) using SequenceMatcher\footnote{\textls[-15]{\url{https://docs.python.org/3/library/difflib.html}}} at threshold 0.9. Through this normalization process, the number of unique tasks was reduced from 3,626 to 3,203.

\paragraph{Evaluation Metric}
In preprocessing, we apply $k$-value normalization to treat metrics with different $k$-values as variants of the same base metric, e.g., BLEU-1,2,4 $\rightarrow$ BLEU; ROUGE-1,2,L $\rightarrow$ ROUGE.

\paragraph{Evaluation Criteria}
We map the 7418 unique criterion strings (combining LLM and human extractions) into the QCET quality taxonomy~\citep{belz-2024-qcet}, a three-level lattice (L1 frame-of-reference $\times$ L2 type $\times$ L3 aspect) with 111 leaf criteria organised under four frame-of-reference dimensions at L1: QO (Quality of outputs in their own right), QI (relative to input), QT (relative to target outputs), and QE (relative to an external frame of reference). Our pipeline runs in three steps. \textbf{(1) Raw criteria to QCET criteria classification.} An LLM classifies each raw variant against the 111 QCET leaves. The model returns either a QCET leaf with a confidence indicator (\textit{strong} or \textit{partial}), or \textit{none} with a short justification. Criteria are processed in batches of 5; \textbf{(2) Clustering to find new leaves.} Criteria with \textit{partial} or \textit{none} match are embedded with \texttt{sentence-transformers/all-MiniLM-L6-v2} and clustered with HDBSCAN, producing 19 thematic clusters plus a noise one with 5,111 criteria. For each of the 19 clusters, an LLM call returns a verdict: keep (construct not captured by any QCET leaf), map (back to a existing leaf covers it), split (cluster mixes distinct criteria to be mapped separately), or drop (extraction noise, not a criterion). Four clusters become new QCET leaves placed at appropriate lattice positions: \textit{Absence of Toxic / Harmful Content} (QOC-c-2), \textit{Engagingness / Interestingness} (QEF-w-10), \textit{Absence of Bias / Stereotypes} (QEC-c-3), and \textit{Empathy / Emotional Appropriateness} (QEF-w-11). We additionally add two leaves based on high frequency in the noise cluster: \textit{Creativity / Non-creativity} (QOF-w-6) and \textit{Refusal Appropriateness} (QEC-w-2). Together these six extensions take QCET from 111 to 117 leaves (Table~\ref{tab:qcet_extensions}). We also introduce two auxiliary categories: \textit{Overall Quality / Preference} for holistic judgements that explicitly refuse decomposition into specific aspects, and \textit{Other / Unclassifiable} for extraction noise (dropped from analysis). \textbf{(3) Re-classification.} Each criterion that were not a strong match in step 1 is mapped against the unified target set of 117 QCET leaves plus the two auxiliary categories.

\begin{table*}[htbp]
\centering
\small
\begin{tabularx}{\textwidth}{@{}l l l c c X@{}}
\toprule
\textbf{New leaf id} & \textbf{Name} & \textbf{Origin} & \textbf{Variants} & \textbf{Paper-occ.} & \textbf{Top raw variants (count)} \\
\midrule
QOC-c-2  & Absence of Toxic / Harmful Content   & cluster   &  95 & 409 & harmfulness (57); safety (53); Safety (48) \\
QEF-w-10 & Engagingness / Interestingness       & cluster   &  48 & 248 & engagingness (36); Engagement (28); interestingness (27) \\
QEC-c-3  & Absence of Bias / Stereotypes        & cluster   & 109 & 172 & fairness (9); Bias (7); bias (7) \\
QEF-w-11 & Empathy / Emotional Appropriateness  & cluster   &  34 & 108 & Empathy (46); empathy (18); Emotional Support (4) \\
QOF-w-6  & Creativity / Non-creativity          & frequency &  26 & 157 & creativity (44); Creativity (39); Novelty (18) \\
QEC-w-2  & Refusal Appropriateness              & frequency &  45 &  63 & Refusal (8); refusal (6); Abstention (2) \\
\bottomrule
\end{tabularx}
\caption{The 6 extension leaves added to QCET (111 $\to$ 117). \textbf{Origin} indicates whether the leaf was discovered by HDBSCAN clustering of stage-1 classification residuals (\textit{cluster}) or by frequency analysis of remaining unclustered residual variants (\textit{frequency}). \textbf{Variants} is the number of distinct raw criterion strings mapped to this leaf; \textbf{Paper-occ.} sums the LaaJ and human paper-occurrences across those variants.}
\label{tab:qcet_extensions}
\end{table*}

\paragraph{Language}
We map the languages according to the list of ISO language codes\footnote{\url{https://en.wikipedia.org/wiki/List_of_ISO_639_language_codes}} and keep the standard ISO language names; however, many multilingual datasets do not explicitly enumerate the investigated languages but instead provide only the designation ``multilingual'' with a language count. 

\paragraph{Model Names} To normalize the models, we preprocess the model names by using a uniform format: family\_version\_size\_extra(name) for all extracted models if applicable. 

\paragraph{Dataset Names} We perform similar preprocessing as other terms: lowercase all letters, replace separators (-,/,\_) with spaces and remove special characters (\&). For WMT datasets, as there are many small variations, we grouped them based on the years of release. 

\subsection{Example of Term Normalization Mapping}
\label{append: term_norm_examples}

We list our term normalization results for the top 10 most frequent terms in Table~\ref{tab:automatic_metrics_merges}-Table~\ref{tab:models_merges}.

\begin{table}[htbp]
\centering
\small
\begin{tabularx}{\columnwidth}{@{}llX@{}}
\toprule
\textbf{Term} & \textbf{Cnt} & \textbf{Top Variants (count)} \\
\midrule
Question Answering & 986 & Question Answering (978); question answering (5); Question-Answering (2) \\
Code Generation & 708 & Code Generation (707); code generation (1) \\
Mathematical Reasoning & 219 & Mathematical Reasoning (213); mathematical reasoning (5); Mathematical reasoning (1) \\
Instruction Following & 211 & Instruction Following (196); Instruction-Following (5); instruction following (3) \\
Language Modeling & 163 & Language Modeling (162); language modeling (1) \\
Data To Text Generation & 136 & Data-to-Text Generation (92); Data-to-Text (36); Data to Text (3) \\
Story Generation & 112 & Story Generation (111); Story generation (1) \\
Explanation Generation & 101 & Explanation Generation (100); explanation generation (1) \\
Semantic Parsing & 88 & Semantic Parsing (86); semantic parsing (2) \\
Commonsense Reasoning & 81 & Commonsense Reasoning (78); commonsense reasoning (2); Commonsense reasoning (1) \\
\bottomrule
\end{tabularx}
\caption{Top-10 normalized tasks and their top variants}
\label{tab:tasks_merges}
\end{table}

\begin{table*}[htbp]
\centering
\small
\begin{tabularx}{\textwidth}{@{}llX@{}}
\toprule
\textbf{Normalized Term} & \textbf{Cnt} & \textbf{Top Variants (count)} \\
\midrule
BLEU & 2384 & BLEU (2380); BLEU-C (1); BLEU-L (1); BLEU-P (1); BLEU@5 (1) \\
Accuracy & 2248 & Accuracy (1938); accuracy (265); Accuracy@1 (8); Accuracy@k (4); Weighted Accuracy (4) \\
ROUGE & 1838 & ROUGE (1774); ROUGE-L (53); ROUGE-2 (3); ROUGE-S (3); ROUGE-1 (2) \\
F1 & 1548 & F1 (1136); F1-score (116); F1 Score (52); Macro-F1 (45); F1 score (28) \\
BERTScore & 886 & BERTScore (880); BERTSCORE (3); BertScore (3) \\
Recall & 796 & Recall (514); recall (54); Recall@k (44); Recall@10 (32); Recall@K (30) \\
Perplexity & 745 & Perplexity (664); perplexity (79); $\Delta$ Perplexity (1); $\epsilon$-perplexity (1) \\
Exact Match & 701 & Exact Match (555); EM (94); exact match (26); Exact match (14); Exact-match (3) \\
Precision & 590 & Precision (447); precision (46); Precision@1 (23); Precision@k (17); Precision@5 (8) \\
CIDEr & 247 & CIDEr (242); CIDER (3); CIDEr-D (1); CIDEr-R (1) \\
\bottomrule
\end{tabularx}
\caption{Top-10 normalized automatic metrics and their top variants}
\label{tab:automatic_metrics_merges}
\end{table*}

\begin{table*}[htbp]
\centering
\small
\begin{tabularx}{\textwidth}{@{}llX@{}}
\toprule
\textbf{QCET Target (id)} & \textbf{Cnt} & \textbf{Top Variants (count)} \\
\midrule
Correctness of Outputs (QOC-w-1) & 480 & correctness (173); accuracy (111); Correctness (83) \\
Factual Truth (QEC-c-1) & 465 & Factuality (36); factuality (32); factual consistency (25) \\
Relevance to Input (QIG-c-3) & 438 & relevance (187); Relevance (100); adequacy (7) \\
Coherence (QOG-c-3) & 279 & coherence (93); Coherence (67); logical coherence (8) \\
Consistency with Input (QIC-c-3) & 260 & faithfulness (43); Faithfulness (23); instruction-following (11) \\
Absence of Toxic Content (QOC-c-2) & 241 & harmfulness (42); harmlessness (35); safety (31) \\
Usefulness (QEG-w-3) & 236 & helpfulness (137); Helpfulness (77); usefulness (9) \\
Overall Quality / Pref. (AUX-OQ) & 223 & overall quality (33); quality (23); Quality (20) \\
Absence of Omissions (QIC-c-1) & 196 & completeness (38); Completeness (36); comprehensiveness (12) \\
Internal Consistency of Outputs (QOG-c-4) & 187 & consistency (56); Consistency (37); logical consistency (9) \\
\bottomrule
\end{tabularx}
\caption{Top-10 LaaJ criteria after QCET-based normalization, with paper count and the most frequent raw variants that mapped to each target.}
\label{tab:llm_criteria_merges}
\end{table*}

\begin{table*}[htbp]
\centering
\small
\begin{tabularx}{\textwidth}{@{}llX@{}}
\toprule
\textbf{QCET Target (id)} & \textbf{Cnt} & \textbf{Top Variants (count)} \\
\midrule
Fluency (QOG-w-3) & 806 & fluency (477); Fluency (269); Language Fluency (6) \\
Relevance to Input (QIG-c-3) & 805 & relevance (280); Relevance (155); adequacy (64) \\
Factual Truth (QEC-c-1) & 778 & factuality (58); Factuality (42); factual consistency (36) \\
Correctness of Outputs (QOC-w-1) & 694 & correctness (211); accuracy (139); Correctness (87) \\
Coherence (QOG-c-3) & 630 & coherence (248); Coherence (163); semantic coherence (8) \\
Overall Quality / Pref. (AUX-OQ) & 499 & overall quality (108); preference (60); quality (46) \\
Consistency with Input (QIC-c-3) & 476 & faithfulness (101); Faithfulness (42); meaning preservation (28) \\
Absence of Omissions (QIC-c-1) & 408 & completeness (65); Completeness (45); comprehensiveness (18) \\
Informativeness (QOG-c-2) & 391 & informativeness (183); Informativeness (97); Informative (5) \\
Internal Consistency of Outputs (QOG-c-4) & 333 & consistency (90); Consistency (71); logical consistency (11) \\
\bottomrule
\end{tabularx}
\caption{Top-10 human evaluation criteria after QCET-based normalization, with paper count and the most frequent raw variants that mapped to each target.}
\label{tab:human_criteria_merges}
\end{table*}

\begin{table}[htbp]
\centering
\small
\begin{tabularx}{\columnwidth}{@{}llX@{}}
\toprule
\textbf{Term} & \textbf{Cnt} & \textbf{Top Variants (count)} \\
\midrule
Chinese & 1142 & Chinese (1106); Mandarin (18); Simplified Chinese (6) \\
German & 1034 & German (1031); Swiss German (1); Swiss-German (1) \\
Spanish & 537 & Spanish (536); spa (1) \\
Russian & 402 & Russian (401); rus (1) \\
Arabic & 322 & Arabic (288); Egyptian Arabic (13); Modern Standard Arabic (7) \\
Portuguese & 219 & Portuguese (210); Brazilian Portuguese (9) \\
Turkish & 219 & Turkish (218); tur (1) \\
Korean & 210 & Korean (209); kor (1) \\
Finnish & 162 & Finnish (161); fin (1) \\
Indonesian & 145 & Indonesian (144); ind (1) \\
\bottomrule
\end{tabularx}
\caption{Top-10 normalized languages and their top variants}
\label{tab:languages_merges}
\end{table}

\begin{table}[htbp]
\centering
\small
\begin{tabularx}{\columnwidth}{@{}llX@{}}
\toprule
\textbf{Term} & \textbf{Cnt} & \textbf{Top Variants (count)} \\
\midrule
GSM8K & 454 & GSM8K (428); GSM8k (18); GSM-8K (6) \\
CNN/DailyMail & 303 & CNN/DailyMail (159); CNN/Daily Mail (70); CNN/DM (51) \\
WMT14 & 289 & WMT14 English-German (40); WMT14 (35); WMT14 En-De (26) \\
XSUM & 230 & XSum (182); XSUM (46); Xsum (1) \\
TriviaQA & 223 & TriviaQA (221); TRIVIAQA (1); triviaQA (1) \\
HotpotQA & 214 & HotpotQA (192); HotPotQA (16); HOTPOTQA (6) \\
HumanEval & 201 & HumanEval (180); HumanEval+ (19); HUMANEVAL (2) \\
\makecell{NATURAL \\ QUESTIONS} & 173 & Natural Questions (172); NATURAL QUESTIONS (1) \\
MATH & 164 & MATH (161); Math (3) \\
SQuAD & 161 & SQuAD (155); SQUAD (3); SQuAD-1 (2) \\
\bottomrule
\end{tabularx}
\caption{Top-10 normalized datasets and their top variants}
\label{tab:datasets_merges}
\end{table}

\begin{table}[htbp]
\centering
\small
\begin{tabularx}{\columnwidth}{@{}llX@{}}
\toprule
\textbf{Term} & \textbf{Cnt} & \textbf{Top Variants (count)} \\
\midrule
GPT-4 & 918 & GPT-4 (916); ChatGPT (GPT-4) (1); SyncTOD (GPT-4) (1) \\
GPT-4o & 905 & GPT-4o (892); GPT-4O (12); gpt-4o (1) \\
Transformer & 601 & Transformer (594); TRANSFORMER (2); transformer (2) \\
BART & 552 & BART (550); GEE (BART) (1); SOV-MAS (BART) (1) \\
GPT-3.5-turbo & 519 & GPT-3.5-turbo (261); GPT-3.5-Turbo (176); GPT-3.5 Turbo (54) \\
LLaMA-2-7B & 507 & LLaMA-2-7B (126); Llama-2-7B (123); LLaMA2-7B (77) \\
GPT-2 & 506 & GPT-2 (493); GPT2 (10); CALM (GPT-2) (2) \\
BERT & 469 & BERT (467); PACSUM (bert) (1); Transformer (BERT) + MLP (1) \\
GPT-3.5 & 427 & GPT-3.5 (422); ChatGPT (GPT-3.5) (4); ChatGPT-API (GPT-3.5) (1) \\
T5 & 427 & T5 (426); SOV-MAS (T5) (1) \\
\bottomrule
\end{tabularx}
\caption{Top-10 normalized NLG models and their top variants}
\label{tab:models_merges}
\end{table}

\subsection{NLG Task Filtering}
\label{append:task_filtering}

Table shows the filtered out tasks among the top-ranked ones (we rank them based on frequency, and keep the top-30 after filtering) and the reason for exclusion.

\begin{table*}[]
    \centering
    \small
    \renewcommand{\arraystretch}{1.2}
    \begin{tabularx}{\textwidth}{@{}lX@{}}
        \toprule
        \textbf{Excluded task} & \textbf{Reason} \\
        \midrule
        Code Generation & Not natural language output; verified by execution, not text quality \\
        Mathematical Reasoning & Evaluated by answer correctness against gold answer, not linguistic criteria \\
        Math Problem Solving & Same as Mathematical Reasoning \\
        Math Reasoning & Same as Mathematical Reasoning \\
        Text-To-SQL Generation & Query output; verified by execution accuracy \\
        Semantic Parsing & Meaning representation; evaluated by exact/graph match \\
        Instruction Following & Measures constraint compliance (task success), not text quality \\
        Language Modeling & Evaluated intrinsically (perplexity) \\
        Natural Language Inference & Classification \\
        Text Classification & Classification \\
        Multiple Choice QA & Discriminative selection scored by accuracy \\
        Sentiment Analysis & Classification \\
        Named Entity Recognition & Sequence labeling scored by span F1; no new text generated \\
        Automatic Speech Recognition & Transcription of given content; not open-ended generation \\
        Text-To-Speech Generation & Audio output; evaluated on acoustic criteria, not text \\
        \bottomrule
    \end{tabularx}
    \caption{Excluded Tasks and Reasons}
    \label{tab:excluded_tasks}
\end{table*}

We further show the top-3 automatic metrics for the top excluded tasks in Table~\ref{tab:new_tasks_metrics}. The top metrics of these tasks confirm our exclusion reasoning: they are dominated by functional-correctness metrics, whereas our included NLG tasks are topped by text-quality metrics.

\begin{table}[htbp]
    \centering
    \small
    \begin{tabularx}{\columnwidth}{@{}XX@{}}
        \toprule
        \textbf{Top excluded tasks} & \textbf{Top-3 evaluation metrics} \\
        \midrule
        Code Generation (401) & Accuracy (33\%), pass@k (30\%), Exact Match (15\%) \\
        Instruction Following (131) & Accuracy (41\%), Win Rate (28\%), ROUGE (16\%) \\
        Math Reasoning, all variants (154) & Accuracy (73\%), pass@k (24\%), Exact Match (14\%) \\
        Text-to-SQL Generation (41) & Execution Accuracy (46\%), Exact Match Accuracy (32\%), Exact Match (24\%) \\
        Semantic Parsing (55) & Exact Match (36\%), Accuracy (29\%), Exact Match Accuracy (24\%) \\
        Language Modeling (102) & Perplexity (78\%), Accuracy (36\%), BLEU (24\%) \\
        \bottomrule
    \end{tabularx}
    \caption{Top-3 metrics for newly added tasks}
    \label{tab:new_tasks_metrics}
\end{table}

\subsection{Human Annotation Details}
\label{append:human}

Details of human-annotation studies referenced in \S\ref{sec:human_annotation} and 4.3 are presented below. The full annotation guidelines are in our project repository.

\paragraph{Annotator pool.}
All 11 annotators are NLP researchers (NLG-related expertise) recruited through the authors' institutions; participation was either by co-authorship or paid student work (see Ethical Statement).

\paragraph{Field-level extraction quality.}
For each of 60 papers (sampled from the 110), an annotator scored each LLM-extracted metadata field on a 1--3 scale:
1 = mismatch (extraction is wrong or substantially incomplete);
2 = partial match (some correct entries, some missing or off);
3 = full match (extraction matches the annotator's reading).
Field-level mean scores are in Table~\ref{tab:likert_extraction}. All 11 fields exceed mean 2.5; LaaJ models, languages, and LaaJ criteria score highest, while NLG models and output formats score lowest.

\begin{table}[htbp]
\centering
\small
\begin{tabular}{@{}lr@{}}
\toprule
\textbf{Field} & \textbf{Mean (1--3)} \\
\midrule
Q1: tasks               & 2.62 \\
Q1: datasets            & 2.71 \\
Q1: languages           & 2.88 \\
Q1: NLG models          & 2.52 \\
Q1: output formats      & 2.52 \\
Q2: automatic metrics   & 2.69 \\
Q3: LaaJ models         & 2.91 \\
Q3: LaaJ methods        & 2.79 \\
Q3: LaaJ criteria       & 2.88 \\
Q4: human-eval methods  & 2.79 \\
Q4: human-eval criteria & 2.79 \\
\bottomrule
\end{tabular}
\caption{LLM-extraction field-level Likert scores (60 papers; 1=mismatch, 2=partial, 3=full match). Means above 2.5 across all fields indicate that LLM extraction substantially aligns with human annotators.}
\label{tab:likert_extraction}
\end{table}

\paragraph{NLG-vs-non-NLG filter validation.}
We randomly sampled 120 papers from the LLM-filtered ``non-NLG'' set. Annotators followed the guideline ``Does the paper address NLG tasks?'' with positive examples (summarization, dialogue, MT, paraphrase, captioning) and negative examples (sentiment analysis, NER, extractive QA without generated text). 6 NLP-researcher annotators each labeled 40 papers, with each paper annotated twice. 117 of 120 papers (97.5\%) had both annotators agreeing on ``No''; the remaining 3 had one ``Yes'' and one ``No''. 

\paragraph{Metric-criterion mapping}
\label{append:mapping}
To check whether papers actually map criteria that their metrics measure, we randomly selected 50 papers which have ``yes'' to Q2 (automatic metrics) from the 3,334 NLG papers. For each paper, we locate the passage where its metrics are introduced and extract for each metric the criterion it was mapped to. Each paper was annotated independently by two annotators (agreement 77.6\%, Cohen’s kappa 0.49).  

Results from manual annotation reveal that out of 183 metrics from the 50 papers, only 28-36\% of the metrics (per annotator; 21\% by both annotators’ consensus) mentioned which criterion it measures; only 6\% of papers stated metric-criterion mapping for every metric used (by consensus; 10-12\% per annotator). 22 of the 31 metrics with a consensus-stated criterion occur only once in our sample, while the most frequently used metrics are rarely mapped (BLEU 12\%, ROUGE 11\%, BERTScore and METEOR 0\%).   
Among the metrics that mentioned the criteria they measure, most are designed for the specific criterion (e.g.: Distinct  for Diversity, Knowledge-F1 for informativeness). Other frequently used metrics have no unique construct: when BLEU, ROUGE or Perplexity are mapped, the stated criteria differ from paper to paper (BLEU: coherence / content retention / relevance; ROUGE: relevance / text reuse / grammar-and-redundancy; Perplexity: fluency / coherence / style). This confirms that general-purpose metrics are widely used as quality estimators without agreed and validated constructs. 

\paragraph{LaaJ-against-Human extraction.}
To evaluate the supplementary LLM extractions for LaaJ-human validation, we sampled 90 papers (72 predicted \texttt{``Yes''} and 18 \texttt{``No''}). Nine NLP researchers annotated them, each reviewing 20 papers to ensure two annotations per paper. For the binary classification (presence of validation), inter-annotator agreement was 75.6\% (Cohen's $\kappa=0.45$). After merging the human annotations by taking the union of extracted rows per paper, the human--LLM agreement reached 88.9\% ($\kappa=0.67$). However, for the per-paper count of distinct extracted rows (metric, criterion, judge-model), the exact match rate was only 21.1\% (Pearson $r=0.32$). Restricting to papers with validation content (LLM: 72; human-union: 70) and deduplicating within (paper, metric, criterion, value), humans extracted an average of 13.86 distinct rows per paper compared to the LLM's 4.31. For the rows the LLM does extract, value fidelity is high: $91\%$ of LLM-extracted values lie within $\pm 0.05$ of a human-annotated value for the same (paper, criterion, metric) cell. The bias is concentrated on aggregate versus fine-grained criteria: for \emph{Overall Quality~/~Preference}, LLM and human extract the same number of rows ($31$ each), but for every other QCET criterion the LLM under-extracts by $3$--$10\times$, with several criteria (\emph{Usefulness for Task}, \emph{Control over Style}, \emph{Absence of Toxic/Harmful Content}) extracted by humans but missed entirely by the LLM. As a share of total extracted rows, $26.3\%$ of LLM-extracted rows are tagged \emph{Overall Quality~/~Preference} versus $5.9\%$ for humans --- a $4.5\times$ over-representation. Manual inspection of the LLM prompts suggests this pattern is caused by the PDF-to-text conversion step in our pipeline rather than the extractor LLM itself: tabular content, where per-criterion correlations are typically reported, is largely lost during PDF parsing, while the text mentions of aggregate scores survive. 

\subsection{Association Measures and LR Formulations}
\label{append: lr_formula}

\paragraph{Likelihood Ratio (LR)}
We list each LR variant below.
For all three, \textit{A} and \textit{B} are sets of papers defined by the presence of a specific term.

Metric (or criteria)--task LR: \textit{A} = papers with a specific task, \textit{B} = papers with a specific metric (note that $B \not\subseteq A$):
\begin{equation}
    LR_{\text{metric}|\text{task}}(A \to B) = \frac{P(B|A)}{P(B|\neg A)}
\end{equation}

Metric--criteria LR: \textit{A} = papers with a specific metric, \textit{B} = papers with a specific criterion:
\begin{equation}
    LR_{\text{criteria}|\text{metric}}(A \to B) = \frac{P(B|A)}{P(B|\neg A)}
\end{equation}

\paragraph{Dunning $G^2$ test}
For a $2{\times}2$ contingency table with cells $(k_{11}, k_{12}, k_{21}, k_{22})$
(where $k_{11}$ = papers with both \textit{A} and \textit{B}, $k_{12}$ = papers with \textit{A} but not \textit{B},
$k_{21}$ = papers with \textit{B} but not \textit{A}, $k_{22}$ = papers with neither),
the $G^2$ statistic is:
\begin{equation}
    G^2 = 2\!\sum_{i,j} O_{ij}\ln\!\frac{O_{ij}}{E_{ij}}, \quad
    E_{ij} = \frac{r_i \cdot c_j}{N}
\end{equation}
where $r_i$ and $c_j$ are row and column marginals and $N$ is the total number of papers.
$G^2 \sim \chi^2_1$ under $H_0$ (independence), giving a $p$-value directly.
We apply Benjamini--Hochberg FDR correction \citep{benjamini1995} to all $p$-values within each analysis family and keep pairs with adjusted $p \leq 0.05$.

\paragraph*{Illustrative example of LR and $G^2$ computation}
Table~\ref{tab:lr_g2_example} illustrates LR and $G^2$ on two real pairs from our data.
In both cases, papers are split by whether they contain term \textit{A} and/or term \textit{B},
forming a $2{\times}2$ contingency table:

\medskip
\begin{center}
\small
\begin{tabular}{lcc}
\toprule
 & \textbf{$B$} & \textbf{$\neg B$} \\
\midrule
\textbf{$A$}   & $k_{11}$ & $k_{12}$ \\
\textbf{$\neg A$} & $k_{21}$ & $k_{22}$ \\
\bottomrule
\end{tabular}
\end{center}
\medskip

\noindent
Example~A is a \emph{task--metric} pair ($A$ = task, $B$ = metric) drawn from all $N=3{,}334$ papers.
Example~B is a \emph{metric--criterion} pair ($A$ = metric, $B$ = criterion) drawn from the
$N=1{,}711$ papers that report human evaluation.
Despite near-identical LR ($\approx9.2$--$9.4$), the two examples reach opposite conclusions because
$k_{11}$ differs by a factor of 45.

\begin{table*}[ht]
\centering
\small
\caption{LR and $G^2$ on two real pairs from our data with nearly the same LR but opposite
  significance outcomes.
  Example~A (task--metric): \textit{A} = question answering, \textit{B} = exact match; full set ($N=3{,}334$).
  Example~B (metric--criterion): \textit{A} = coverage, \textit{B} = relative factual accuracy (QCET QEC-c-2);
  human-evaluation subset ($N=1{,}711$).
  The large difference in $k_{11}$ (136 vs.\ 3) drives the difference in $E_{11}$ and hence $G^2$,
  producing opposite conclusions after BH-FDR correction.}
\label{tab:lr_g2_example}
\begin{tabular}{lrr}
\toprule
 & \textbf{Example A} & \textbf{Example B} \\
 & \textit{(QA $\to$ exact match)} & \textit{(coverage $\to$ rel.\ factual accuracy)} \\
\midrule
$N$ (papers)                                 & 3{,}334 & 1{,}711 \\
$k_{11}$ (\textit{A} and \textit{B})         & 136     & 3 \\
$k_{12}$ (\textit{A}, not \textit{B})        & 326     & 19 \\
$k_{21}$ (\textit{B}, not \textit{A})        & 90      & 25 \\
$k_{22}$ (neither)                           & 2{,}782 & 1{,}664 \\
\midrule
$P(B|A)=k_{11}/n_+$                          & $136/462=0.294$ & $3/22=0.136$ \\
$P(B|\neg A)=k_{21}/n_-$                     & $90/2872=0.031$ & $25/1689=0.015$ \\
$\mathrm{LR}=P(B|A)/P(B|\neg A)$             & \textbf{9.4}    & \textbf{9.2} \\
\midrule
$E_{11}=n_+(k_{11}+k_{21})/N$                & 31.3            & 0.36 \\
$G^2$                                         & 292.4           & 8.0 \\
Raw $p$-value                                 & ${<}0.001$      & $0.0046$ \\
BH-FDR-adj.\ $p$                              & ${<}0.001$      & $0.175$ \\
\midrule
Significant (adj.\ $p \leq 0.05$)?            & \textbf{Yes}    & \textbf{No} \\
\bottomrule
\end{tabular}
\end{table*}


\clearpage 

\section{Additional Results}

\label{append:results} 

\begin{table}[!ht]
    \centering
    \footnotesize 
    \setlength{\tabcolsep}{3pt} 
    \begin{tabularx}{\columnwidth}{l X}
        \toprule
        \textbf{Key} & \textbf{Description} \\
        \midrule
        \rowcolor{gray!10} \multicolumn{2}{l}{\textbf{Q1: NLG task presence (\texttt{answer\_1})}} \\
        \texttt{answer} & ``Yes'' if the paper addresses any NLG task. \\
        \texttt{quote} & Verbatim excerpt supporting the decision. \\
        \texttt{tasks} & NLG tasks (e.g., Summarization, MT, Paraphrase Gen, or \texttt{Other:\textless task\textgreater}). \\
        \texttt{datasets} & List of datasets used for generation/evaluation. \\
        \texttt{languages} & List of languages (e.g., English, Chinese). \\
        \texttt{models} & List of models used to generate outputs. \\
        \texttt{outputs} & Short description of the generated output type. \\
        \midrule
        \rowcolor{gray!10} \multicolumn{2}{l}{\textbf{Q2: Automatic evaluation metrics (\texttt{answer\_2})}} \\
        \texttt{answer} & ``Yes'' if automatic metrics are used. \\
        \texttt{quote} & Excerpt mentioning automatic evaluation. \\
        \texttt{metrics} & List of metrics (e.g., BLEU, BERTScore). \\
        \midrule
        \rowcolor{gray!10} \multicolumn{2}{l}{\textbf{Q3: LaaJ (\texttt{answer\_3})}} \\
        \texttt{answer} & ``Yes'' if an LLM is used \emph{after} generation. \\
        \texttt{quote} & Excerpt describing LLM-based evaluation. \\
        \texttt{models} & Names of LLMs used as judges. \\
        \texttt{methods} & Procedure (e.g., pairwise, scoring prompt). \\
        \texttt{criteria} & Rubric properties (e.g., fluency, coherence). \\
        \midrule
        \rowcolor{gray!10} \multicolumn{2}{l}{\textbf{Q4: Human evaluation (\texttt{answer\_4})}} \\
        \texttt{answer} & ``Yes'' if humans evaluate generated outputs. \\
        \texttt{quote} & Excerpt mentioning human evaluation. \\
        \texttt{guideline} & Instructions given to human raters. \\
        \texttt{criteria} & Explicit criteria (e.g., fluency, coherence). \\
        \bottomrule
    \end{tabularx}
    \caption{Schema of the JSON object returned by the LLM for each paper.}
    \label{tab:llm-schema}
    \vspace{15pt}
\end{table}

\begin{table}[ht]
    \begin{adjustbox}{width=\columnwidth}
    \centering
    \setlength{\tabcolsep}{6.2pt}
    \begin{tabular}{lcccc}
        \toprule
        \textbf{Model} & \textbf{Q1} & \textbf{Q2} & \textbf{Q3} & \textbf{Q4} \\
        \midrule
        DeepSeek-R1 & 91.97 & 93.69 & 95.28 & 94.32 \\
        GPT-OSS-120B & 80.68 & 89.86 & 95.52 & 94.93 \\
        Qwen3-235B & 92.42 & 94.58 & 95.01 & 95.14 \\
        \midrule
        Krippendorff's $\alpha$ & 0.7101 & 0.6879 & 0.8048 & 0.8124 \\
        \bottomrule
    \end{tabular}
    \end{adjustbox}
    \caption{Pairwise agreement (\%) between each LLM and LLM-harmonized results (row 1-3), and Krippendorff's $\alpha$ among the three LLMs.}
    \label{tab:pairwise-agreement}
    \vspace{15pt}
\end{table}

\begin{table}[h]
\small
\begin{adjustbox}{width=\columnwidth}
  \centering
  \setlength{\tabcolsep}{8.6pt}
  \begin{tabular}{lcccc}
    \toprule
    \textbf{Model} & \textbf{Q1} & \textbf{Q2} & \textbf{Q3} & \textbf{Q4} \\
    \midrule
    DeepSeek-R1 & 63.5 & 69.0 & 16.0 & 27.1 \\
    GPT-OSS-120B & 48.5 & 59.9 & 15.4 & 26.7 \\
    Qwen3-235B & 65.1 & 70.3 & 18.7 & 34.4 \\
    Majority Voting & 60.8 & 65.9 & 15.9 & 28.3 \\
    LLM-harmonized & 56.6 & 56.5 & 14.0 & 26.5 \\
    \bottomrule
  \end{tabular}
  \end{adjustbox}
  \caption{\textit{Yes} percentage across all models and questions, for the four binary questions from three LLMs, along with their majority votes and LLM-harmonized results.}
  \label{tab:overall-summary}
  \vspace*{-1em}
\end{table}

\begin{table*}[!htb]
    \centering
    \renewcommand*{\arraystretch}{0.875}
    \begin{adjustbox}{width=\textwidth}
    \setlength{\tabcolsep}{7pt}
    \begin{tabular}{lccccccccccc}
        \toprule
        \textbf{Conference} & \textbf{Total} & \textbf{NLG} & \textbf{NLG\%} & \textbf{Tasks} & \textbf{Datasets} & \textbf{Languages} & \textbf{NLG Models} & \textbf{Auto Metrics} & \textbf{LaaJ Crit.} & \textbf{Human Crit.} & \textbf{LaaJ Models} \\
        \midrule
        ACL-2020 & 778 & 297 & 38.2 & 115 & 561 & 92 & 1029 & 314 & 0 & 227 & 0 \\
        ACL-2021 & 571 & 243 & 42.6 & 124 & 582 & 76 & 751 & 277 & 0 & 231 & 1 \\
        ACL-2022 & 603 & 275 & 45.6 & 196 & 686 & 88 & 842 & 365 & 5 & 284 & 5 \\
        ACL-2023 & 910 & 463 & 50.9 & 319 & 1202 & 162 & 1211 & 582 & 28 & 409 & 22 \\
        ACL-2024 & 864 & 591 & 68.4 & 506 & 1546 & 190 & 1392 & 841 & 375 & 586 & 97 \\
        ACL-2025 & 1600 & 1096 & 68.5 & 832 & 2596 & 233 & 2221 & 1667 & 918 & 1050 & 309 \\
        \midrule
        EMNLP-2020 & 751 & 291 & 38.7 & 159 & 578 & 86 & 910 & 323 & 0 & 219 & 1 \\
        EMNLP-2021 & 847 & 326 & 38.5 & 151 & 674 & 131 & 1004 & 410 & 1 & 270 & 4 \\
        EMNLP-2022 & 828 & 381 & 46.0 & 247 & 950 & 193 & 1062 & 436 & 14 & 358 & 23 \\
        EMNLP-2023 & 1047 & 553 & 52.8 & 395 & 1401 & 231 & 1297 & 727 & 148 & 477 & 37 \\
        EMNLP-2024 & 1268 & 823 & 64.9 & 634 & 1985 & 339 & 1735 & 1122 & 506 & 706 & 160 \\
        EMNLP-2025 & 1809 & 1374 & 76.0 & 994 & 3073 & 225 & 2438 & 2027 & 1064 & 1123 & 341 \\
        \midrule
        INLG-2020 & 46 & 44 & 95.7 & 31 & 96 & 11 & 137 & 78 & 3 & 56 & 1 \\
        INLG-2021 & 45 & 44 & 97.8 & 29 & 89 & 28 & 116 & 72 & 0 & 89 & 0 \\
        INLG-2022 & 25 & 23 & 92.0 & 24 & 57 & 12 & 89 & 44 & 0 & 47 & 0 \\
        INLG-2023 & 36 & 36 & 100.0 & 27 & 94 & 23 & 107 & 77 & 9 & 65 & 2 \\
        INLG-2024 & 53 & 51 & 96.2 & 43 & 100 & 28 & 212 & 103 & 32 & 86 & 7 \\
        INLG-2025 & 50 & 43 & 86.0 & 36 & 121 & 15 & 146 & 101 & 58 & 75 & 41 \\
        \midrule
        NAACL-2021 & 477 & 177 & 37.1 & 105 & 394 & 55 & 554 & 279 & 1 & 171 & 2 \\
        NAACL-2022 & 442 & 184 & 41.6 & 125 & 462 & 62 & 569 & 262 & 0 & 191 & 0 \\
        NAACL-2024 & 486 & 288 & 59.3 & 253 & 850 & 234 & 740 & 445 & 144 & 313 & 42 \\
        NAACL-2025 & 635 & 416 & 65.5 & 331 & 1131 & 153 & 1023 & 770 & 400 & 468 & 114 \\
        \midrule
        \textbf{Total} & \textbf{14171} & \textbf{8019} & \textbf{56.6} & \textbf{3308} & \textbf{11289} & \textbf{801} & \textbf{12995} & \textbf{6805} & \textbf{2752} & \textbf{4953} & \textbf{783} \\
        \bottomrule
    \end{tabular}
    \end{adjustbox}
    \caption{Statistics of total number of papers, NLG papers, and unique number of term counts of tasks, datasets, languages, NLG models, automatic metrics, LaaJ criteria, human criteria, and LaaJ models. NLG papers are counted after LLM harmonization. The number of total papers in ACL and EMNLP increased significantly in 2025.}
    \label{tab:conference_stats}
\end{table*}

\clearpage
\begin{figure*}[t]
    \centering
    \includegraphics[width=\linewidth]{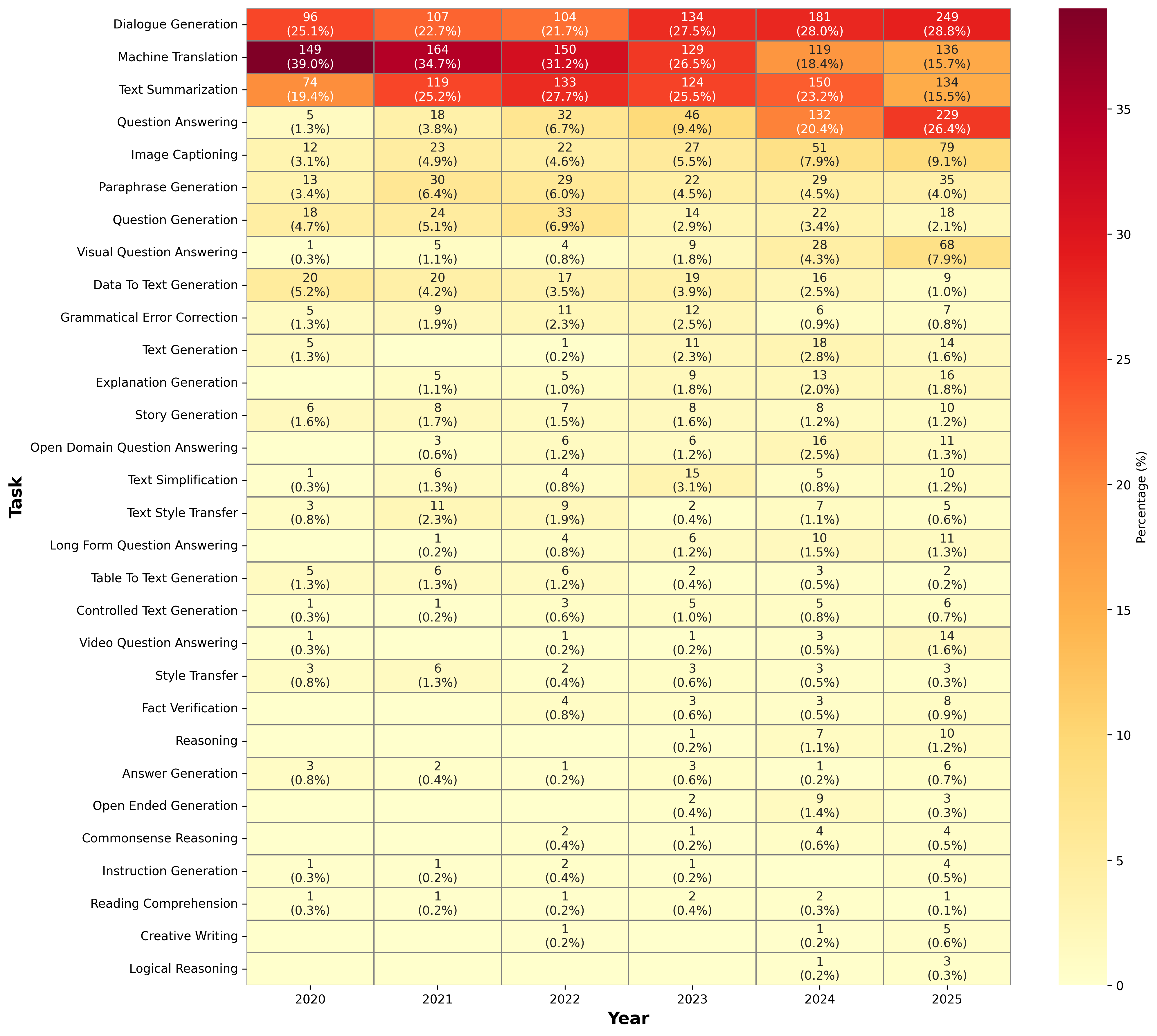}
    \caption{Heatmap of task-year distribution of the top-30 tasks. Both the total number and percentage per year are shown. The percentage of MT is decreasing while QA-related tasks increase significantly (QA, VQA).}
    \label{fig:task_top}
    \vspace*{-1em}
\end{figure*}

\clearpage
\begin{figure*}[t]
    \centering
    \includegraphics[width=\linewidth]{imgs/frequency_by_rank_dashboard.png}
    \caption{Distribution of metadata by tasks, columns include top-four and all-tasks, and rows are the metadata extracted (counts are after term normalization).}
    \label{fig:task_dashboard}
    \vspace*{-1em}
\end{figure*}

\clearpage

\section{Task-Specific Evaluation Guidelines}
\label{append:recommendations}

We distill our findings~(\S\ref{sec:results}) into concrete, criterion-level evaluation recommendations for each of the four major NLG tasks (Table~\ref{tab:task_guidance}).
For each task--criterion pair, we list metrics whose strong task or criterion association in our data, supported by external validation studies, makes them well-suited for that purpose.
We also list metrics to \emph{avoid} as primary evidence for a given criterion---those that co-occur frequently but exhibit low LR with the target criterion, indicating indiscriminate application rather than deliberate, validated selection.
These recommendations are grounded in the trends documented in Figures~\ref{fig:task_four}--\ref{fig:laaj_human} and are intended as a starting point; readers should consult the cited references for task-specific validation evidence before adopting any metric.

\paragraph{Minimal Evaluation Recipes per Task.}
Building on Table~\ref{tab:task_guidance}, we distill a minimal three-to-four-step protocol for each task that researchers can adopt or adapt:

\begin{itemize}[nosep,labelsep=0.2em,align=parleft,topsep=0pt,partopsep=0pt]
     \item \textbf{MT:} (1)~Report COMET~\cite{rei-2020-comet} (or the emerging MetricX series~\cite{juraska-2025-metricx}) as the primary adequacy metric; (2)~retain BLEU only for comparability with prior work~\cite{mathur-2020-tangled-up-in-bleu}, not as a standalone quality signal; (3)~if LaaJ is deployed for MT-specific criteria (e.g., terminology, style), validate it against expert human MT-error annotations on the same test set.
    \item \textbf{DG:} (1)~Report Distinct-$n$~\cite{li-2016-diversity-promoting} for diversity; (2)~deploy LaaJ with a multi-criterion rubric covering engagement (appropriateness, naturalness) and safety (harmlessness, helpfulness); (3)~reserve human evaluation for the safety dimension at minimum, since it is the criterion where LaaJ uniquely surfaces categories that traditional human evaluation rarely covers (\S\ref{sec:results}).
    \item \textbf{TS} (1)~Report ROUGE as a coverage baseline; (2)~select at least one criterion-specific metric matched to your primary quality concern: SummaC~\cite{laban-2022-summac} or UniEval~\cite{zhong-2022-unieval} for consistency, QAFactEval~\cite{fabbri-2022-qafacteval} or MiniCheck~\cite{tang-2024-minicheck} for factuality, MoverScore~\cite{zhao-2019-moverscore} or BARTScore~\cite{yuan-2021-bartscore} for coherence --- noting that adoption of these criterion-specific metrics remains a minority of TS papers in our corpus, so reporting them alongside ROUGE (rather than in place of it) is the realistic transition path; (3)~if factuality is critical, pair automatic metrics with LaaJ factuality assessment on a human-validated subset.
    \item \textbf{QA:} (1)~Report EM and F$_1$ only for comparability with prior work, not as primary signals for open-ended QA; (2)~for open-ended or free-form answers, add a semantic evaluation layer --- LaaJ with a correctness rubric is the most widely adopted option in our corpus, with emerging alternatives such as AlignScore~\cite{zha-2023-alignscore}; (3)~if factual accuracy matters, supplement with an emerging fact-verification metric such as MiniCheck~\cite{tang-2024-minicheck}; (4)~document the expected limitations of each metric (e.g., ``EM ignores paraphrase; BERTScore captures semantic overlap but not factual correctness'').
\end{itemize}

These recipes are intentionally minimal: they define a starting set rather than a comprehensive evaluation suite, and each step can be validated against the cited literature before deployment.

\begin{table*}[htbp]
    \centering
    \small
    \setlength{\tabcolsep}{4pt}
    \renewcommand{\arraystretch}{1.15}
    \begin{tabularx}{\textwidth}{@{}>{\raggedright\arraybackslash}p{2.4cm} >{\raggedright\arraybackslash}p{2.0cm} >{\raggedright\arraybackslash}p{3.0cm} >{\raggedright\arraybackslash}p{2.6cm} >{\raggedright\arraybackslash}p{2.1cm} >{\raggedright\arraybackslash}p{2.0cm}@{}}
    \toprule
    \textbf{Task} & \textbf{Criterion} & \textbf{Recommended Metric(s)} & \textbf{Avoid as Primary} & \textbf{LaaJ viable?} & \textbf{Key References} \\
    \midrule
    \textbf{Machine Translation}
        & Adequacy / Translation Quality & COMET~\cite{rei-2020-comet}, MetricX-25~\cite{juraska-2025-metricx} (COMET 39.0\% of 2025 MT; BLEU still 61.8\%) & BLEU (task-LR$=2.63$, low criterion-LR) & \ding{51} (system-level; segment-level weaker~\cite{lavie-2025-wmt-eval}) & \cite{lavie-2025-wmt-eval, mathur-2020-tangled-up-in-bleu} \\
    \midrule
    \multirow{3}{2.4cm}{\textbf{Dialogue Generation}}
    & Diversity & Distinct-$n$, Self-BLEU~\cite{li-2016-diversity-promoting} & BLEU (task-LR$\approx 1$ for DG) & \ding{55} (auto metrics sufficient) & \cite{li-2016-diversity-promoting} \\
    \cmidrule{2-6}
    & Engagement / Appropriateness & LaaJ + human (multi-item Likert) & ROUGE (n-gram overlap) & \ding{51} (rubric) & \cite{van-der-lee-etal-2019-best, bavaresco-etal-2025-llms} \\
    \cmidrule{2-6}
    & Safety / Harmlessness & LaaJ rubric + human red-teaming (47 DG papers use LaaJ with safety criteria) & --- & \ding{51} (validate vs.\ humans; judge: GPT-4/GPT-4o) & \cite{bavaresco-etal-2025-llms, wang-2024-llms-are-not-fair-evaluators} \\
    \midrule
    \multirow{4}{2.4cm}{\textbf{Text Summarization}}
        & Consistency / Faithfulness (vs.\ input) & SummaC~\cite{laban-2022-summac} (28), UniEval~\cite{zhong-2022-unieval} (23) & ROUGE (low criterion-LR with consistency) & \ding{51} (validation evidence thin, \S\ref{sec:laaj_human_subsec}) & \cite{laban-2022-summac, zhong-2022-unieval} \\
    \cmidrule{2-6}
        & Factual Accuracy & QAFactEval~\cite{fabbri-2022-qafacteval} (12), MiniCheck~\cite{tang-2024-minicheck} (7, mostly 2024--2025) & ROUGE / BERTScore (similarity $\neq$ factuality) & \ding{51} (validation required, \S\ref{sec:laaj_human_subsec}) & \cite{fabbri-2022-qafacteval, tang-2024-minicheck} \\
    \cmidrule{2-6}
    & Coherence / Redundancy & MoverScore~\cite{zhao-2019-moverscore}, BARTScore~\cite{yuan-2021-bartscore} & BLEU (task-LR$=0.54$ in TS) & \ding{55} (well-validated) & \cite{yuan-2021-bartscore, zhao-2019-moverscore} \\
    \cmidrule{2-6}
    & Coverage / Informativeness & LaaJ rubric (coverage scoring), ROUGE (as secondary) & --- & \ding{51} (rubric) & \cite{belz-etal-2025-standard} \\
    \midrule
    \multirow{2}{2.4cm}{\textbf{Question Answering}}
        & Semantic Correctness & LaaJ (pairwise / rubric), AlignScore~\cite{zha-2023-alignscore} & EM / F$_1$ (ignore paraphrase; EM still 27.1\% of 2025 QA) & \ding{51} (primary for open-ended QA; 137/462 QA papers adopt LaaJ) & \cite{bavaresco-etal-2025-llms, zha-2023-alignscore} \\
    \cmidrule{2-6}
        & Factual Accuracy & MiniCheck~\cite{tang-2024-minicheck}, AlignScore~\cite{zha-2023-alignscore} & BERTScore (semantic sim.\ $\neq$ factuality) & \ding{51} (validation required, \S\ref{sec:laaj_human_subsec}) & \cite{tang-2024-minicheck, zha-2023-alignscore} \\
    \bottomrule
    \end{tabularx}
    \caption{Task-specific evaluation guidance derived from our empirical analysis (\S\ref{sec:results}). ``Avoid as Primary'' lists metrics that co-occur frequently but exhibit \emph{low} LR with the target criterion, indicating inertial use rather than validated selection. \ding{51} and \ding{55} indicate whether an LLM-as-a-judge evaluator is recommended as a viable complement. All recommendations should be validated on the specific task and language before deployment.}
    \label{tab:task_guidance}
\end{table*}

\paragraph{How to read this table.}
Table~\ref{tab:task_guidance} should be read row-wise: for each task, we identify the key quality criteria that our data show are both (a) frequently evaluated and (b) poorly served by generic metrics. For \textbf{Machine Translation}, BLEU appears in 81.5\% of MT papers overall and declines from 88.6\% (2020) to 61.8\% (2025), while COMET grows to 39.0\% adoption in 2025; xCOMET grew from 0 papers in 2023 to 18 in 2025, and MetricX-25 from 0 to 12. We therefore recommend COMET and MetricX-25 as primary adequacy metrics and suggest retaining BLEU only for historical comparison~\cite{mathur-2020-tangled-up-in-bleu}. For \textbf{Dialogue Generation}, LaaJ uniquely surfaces safety criteria (\textit{Harmlessness}/\textit{Safety}/\textit{Non-Toxicity}) that traditional human evaluation rarely covers (\S\ref{sec:results}); 47 papers in our corpus deploy LaaJ with explicit safety criteria, predominantly using GPT-4/GPT-4o as the judge. For \textbf{Text Summarization}, UniEval (23 papers) and SummaC (28 papers) are the most commonly adopted specialized factuality metrics in our corpus, with QAFactEval (12 papers) and MiniCheck (7 papers, concentrated in 2024--2025) as growing alternatives. For \textbf{Question Answering}, where Exact Match still anchors 27.1\% of 2025 papers despite the free-form nature of modern LLM-generated answers, 137 of 462 QA papers in our top-30 corpus use LaaJ-based evaluation, and AlignScore appears in 19 papers across the corpus (14 in TS, 6 in QA). We stress that LaaJ recommendations are conditional on human validation: our data show fewer than 8\% of NLG papers validate LaaJ against human judgements, and where validation does exist, per-criterion alignment is highly variable (\S\ref{sec:laaj_human_subsec}).

%
%


\tcbset{
    promptlisting/.style={
        enhanced,
        listing engine=listings,
        listing only,
        listing options={
            basicstyle=\ttfamily\small,
            breaklines=true,
            breakatwhitespace=false,
            breakindent=0pt,
            postbreak=\mbox{\textcolor{gray}{$\hookrightarrow$}\space},
            columns=fullflexible,
            keepspaces=true,
            showstringspaces=false,
            extendedchars=true
        },
        width=\linewidth,
        colback=gray!5,
        colframe=gray!75,
        boxrule=0.5pt,
        arc=0pt,
        left=3pt,
        right=3pt,
        top=5pt,
        bottom=5pt,
        breakable,
        before skip=10pt,
        after skip=10pt,
        pad at break*=2mm
    }
}

\section{Prompts for annotation}
\subsection{Prompt 1: Initial Extraction}
\label{appendix:prompt1}

\begin{tcblisting}{promptlisting}
You are an expert NLP researcher with deep experience in Natural-Language Generation (NLG).

TASK
Read the paper provided below and answer the four numbered questions. Return **only** a single, valid JSON object (no markdown, no comments, no trailing commas).

PAPER
{full_paper_text}

QUESTIONS

1. Does the paper address NLG tasks?
2. Does the paper use automatic metrics to evaluate the generated outputs?
3. Does the paper use Large-Language Models (LLMs) as judges
   (i.e., *after* generation, an LLM is used to judge/assess the outputs)?
4. Does the paper conduct *human* evaluations of the generated outputs?

ANSWER FORMAT (strict)

{
  "answer_1": {
    "answer": "Yes|No",
    "quote": "...",
    "tasks": ["Text Summarization", "Machine Translation", "Other:<task>"],
    "datasets": ["..."],
    "languages": ["English","Chinese","German","..."],
    "models": ["..."],
    "outputs": "..."
  },
  "answer_2": {
    "answer": "Yes|No",
    "quote": "...",
    "automatic_metrics": ["..."]
  },
  "answer_3": {
    "answer": "Yes|No",
    "quote": "...",
    "models": ["..."],
    "methods": ["pairwise evaluation", "..."]
    "criteria": ["fluency","coherence","..."]
  },
  "answer_4": {
    "answer": "Yes|No",
    "quote": "...",
    "guideline": "...",
    "criteria": ["fluency", "coherence", "..."]
  }
}

INSTRUCTIONS & CONSTRAINTS

* If the answer is "No", set all other fields in that section to an empty string ("") or an empty list ([]).

* For **answer_1.tasks** choose one or more from: {"Text Summarization","Dialogue Generation","Paraphrase Generation", "Machine Translation","Image Captioning","Code Generation"}. If none apply, use "Other:<task name>".

* **Answer-2 guidance (automatic evaluation metrics)**
* The **automatic_metrics** must be a list of automatic metrics used to evaluate the generated outputs.

* **Answer-3 guidance (LLM as judge)**
  1. Answer **Yes** only if an LLM is used *after* generation to assess the outputs.
  2. **methods** - short name/description of the evaluation procedure or prompt.
  3. **criteria** - list the rubric properties the LLM is asked to score (e.g., "fluency","relevance","helpfulness"). If the prompt does not specify criteria, leave as an empty list [].

* **Answer-4 guidance (human evaluation)**
  * The **quote** must mention humans, annotators, raters, a crowdsourcing platform, or a similar human-evaluation indicator.
  * The **guideline** must mention questions or criteria for the evaluation.
  * The **criteria** must be explicitly mentioned in the human evaluation, list all criteria. If the paper does not specify criteria, leave as an empty list [].

* The **quote** fields must be verbatim excerpts from the paper (use ellipses ... to shorten if needed).
* Use double quotes for all JSON strings; do **not** use backticks.
* Do not add any keys, text, or formatting other than the JSON object.
\end{tcblisting}

\subsection{Prompt 2: Verification and Normalization}
\label{appendix:prompt2}

\begin{tcblisting}{promptlisting}
You are verifying and improving the extracted metadata from a research paper about natural language generation (NLG) evaluation. Your task is to:

1. **Verify** the extracted yes/no answers are correct
2. **Normalize** metadata to use canonical forms (e.g., "BLEU" instead of "bleu")
3. **Correct** any incorrect items
4. **Add** any missing important items
5. **Remove** any irrelevant or incorrect items

## Paper Information

**Paper ID:** {paper_id}
**Title:** {title}
**Abstract:** {abstract}

**Full Paper Text:**
{full_text}

---

## Extracted Metadata to Review

### Question 1: Does the paper address NLG tasks?

**Extracted Answer:** {answer_1_answer}
**Extracted Metadata:**
- **Tasks:** {answer_1_tasks}
- **Datasets:** {answer_1_datasets}
- **Languages:** {answer_1_languages}
- **Models:** {answer_1_models}
- **Outputs:** {answer_1_outputs}

---

### Question 2: Does the paper use automatic metrics to evaluate the generated outputs?

**Extracted Answer:** {answer_2_answer}
**Extracted Metadata:**
- **Automatic Metrics:** {answer_2_metrics}

---

### Question 3: Does the paper use Large-Language Models (LLMs) as judges (i.e., *after* generation, an LLM is used to judge/assess the outputs)?

**Extracted Answer:** {answer_3_answer}
**Extracted Metadata:**
- **Models:** {answer_3_models}
- **Methods:** {answer_3_methods}
- **Criteria:** {answer_3_criteria}

---

### Question 4: Does the paper conduct *human* evaluations of the generated outputs?

**Extracted Answer:** {answer_4_answer}
**Extracted Metadata:**
- **Guideline:** {answer_4_guideline}
- **Criteria:** {answer_4_criteria}

---

## Your Task

For each question above:
1. **Verify the Yes/No answer** - Is it correct based on the full paper text?
2. **Review the metadata lists** - For each item:
   - Is it correctly extracted from the paper?
   - Is it relevant to the specific question?
   - Should it be normalized? (e.g., "BLEU" vs "bleu", "GPT-3" vs "gpt-3")
3. **Add missing items** - Are there important items mentioned in the paper that are missing?
4. **Remove incorrect items** - Are there items that shouldn't be there?

## Guidelines

### Normalization Rules
- Use canonical/standard forms (e.g., "BLEU" not "bleu", "GPT-3" not "gpt-3")
- Use consistent capitalization for metrics, models
- Use title case for tasks (e.g., "Machine Translation")
- **For metrics**: Simplify to base form (e.g., "BLEU-1", "BLEU-2", "BLEU-4" -> all become "BLEU"; "ROUGE-1", "ROUGE-2", "ROUGE-L" -> all become "ROUGE")
- **For models**: Keep version numbers distinct (e.g., "GPT-3", "GPT-4", "BERT-base", "BERT-large" are different)
- Merge case variations and abbreviations that refer to the same thing

### Verification Rules
- Only include items **explicitly mentioned** in the paper
- Focus on the **main contributions** - don't include every model/dataset mentioned in passing
- For tasks: Only include NLG tasks that are actually evaluated/studied
- For metrics: Include all automatic metrics used to evaluate NLG outputs
- For criteria: Include evaluation criteria used for human eval or LLM-as-evaluator
- Be accurate over complete - it's better to miss minor details than include wrong information

### Answer-Specific Guidelines

**Question 1 (Does the paper address NLG tasks?):**
- Answer "Yes" only if the paper studies/evaluates natural language GENERATION (not just understanding/classification)
- **Tasks**: Choose from {"Text Summarization", "Dialogue Generation", "Paraphrase Generation", "Machine Translation", "Image Captioning", "Code Generation"}. If none apply, use "Other:<task name>".
- **Datasets**: NLG datasets used
- **Languages**: Languages of the generated outputs (e.g., "English", "Chinese", "German")
- **Models**: NLG models being evaluated
- **Outputs**: Description of what is being generated

**Question 2 (Does the paper use automatic metrics to evaluate the generated outputs?):**
- Answer "Yes" if the paper uses any automatic metrics to evaluate generated text
- **Automatic Metrics**: List of automatic evaluation metrics (e.g., BLEU, ROUGE, METEOR, BERTScore)
- **Important**: Simplify metric variants to base form

**Question 3 (Does the paper use LLMs as judges?):**
- Answer "Yes" only if an LLM is used *after* generation to assess the outputs (not just as generation model)
- **Models**: Which LLMs are used for evaluation (e.g., "GPT-4", "Claude-3")
- **Methods**: Short name/description of the evaluation procedure or prompt (e.g., "pairwise evaluation", "direct scoring")
- **Criteria**: List the rubric properties the LLM is asked to score (e.g., "fluency", "relevance", "helpfulness"). If not specified, use empty list.

**Question 4 (Does the paper conduct human evaluations?):**
- Answer "Yes" if humans, annotators, raters, or crowdsourcing are used to evaluate generated outputs
- **Guideline**: Description of questions or criteria for the evaluation
- **Criteria**: List all criteria explicitly mentioned (e.g., "fluency", "coherence", "relevance"). If not specified, use empty list.

## Output Format

Please return ONLY a JSON object with the following structure:
{
  "paper_id": "{paper_id}",
  "answer_1": {
    "answer": "Yes"/"No",
    "answer_changed": true/false,
    "tasks": ["normalized_task1", ...],
    "datasets": ["normalized_dataset1", ...],
    "languages": ["normalized_language1", ...],
    "models": ["normalized_model1", ...],
    "outputs": ["output_description1", ...],
    "changes_made": {
      "added": {"tasks": [...], "datasets": [...], ...},
      "removed": {"tasks": [...], "datasets": [...], ...},
      "normalized": {"original_item": "normalized_item", ...},
      "explanation": "Brief explanation of major changes"
    }
  },
  "answer_2": {
    "answer": "Yes"/"No",
    "answer_changed": true/false,
    "automatic_metrics": ["normalized_metric1", ...],
    "changes_made": {
      "added": {"automatic_metrics": [...]},
      "removed": {"automatic_metrics": [...]},
      "normalized": {"original_metric": "normalized_metric", ...},
      "explanation": "Brief explanation of major changes"
    }
  },
  "answer_3": {
    "answer": "Yes"/"No",
    "answer_changed": true/false,
    "models": ["normalized_model1", ...],
    "methods": ["normalized_method1", ...],
    "criteria": ["normalized_criterion1", ...],
    "changes_made": {
      "added": {"models": [...], "methods": [...], "criteria": [...]},
      "removed": {"models": [...], "methods": [...], "criteria": [...]},
      "normalized": {"original_item": "normalized_item", ...},
      "explanation": "Brief explanation of major changes"
    }
  },
  "answer_4": {
    "answer": "Yes"/"No",
    "answer_changed": true/false,
    "guideline": ["guideline1", ...],
    "criteria": ["normalized_criterion1", ...],
    "changes_made": {
      "added": {"guideline": [...], "criteria": [...]},
      "removed": {"guideline": [...], "criteria": [...]},
      "normalized": {"original_item": "normalized_item", ...},
      "explanation": "Brief explanation of major changes"
    }
  },
  "overall_notes": "Any general observations"
}

## Important Notes

- **Be conservative with changes** - Only modify if you're confident
- **Prioritize accuracy** - Better to keep existing correct items than to add uncertain ones
- **Normalize consistently** - Use standard naming conventions
- **Document major changes** - Explain why you added/removed important items
- **Use the full paper text** - Read the complete paper to verify all metadata is accurate and complete
\end{tcblisting}

\subsection{Prompt 3: LLM-Human Validation Extraction}
\label{appendix:prompt3}

\begin{tcblisting}{promptlisting}
You are analyzing a research paper that uses **both** LLM-as-a-judge (LLM evaluators) and human evaluation to assess natural language generation outputs. Your task is to extract detailed information about how (or whether) the paper validates LLM evaluation against human evaluation.

## Paper Information

**Paper ID:** {paper_id}
**Title:** {title}
**Abstract:** {abstract}
**Full Paper Text:** {full_text}

---

## Previously Extracted Metadata

### LLM-as-a-Judge (Answer 3)
- **Models:** {answer_3_models}
- **Methods:** {answer_3_methods}
- **Criteria:** {answer_3_criteria}

### Human Evaluation (Answer 4)
- **Guideline:** {answer_4_guideline}
- **Criteria:** {answer_4_criteria}

---

## Your Task

Extract information about **validation** of LLM-as-a-judge against human evaluation. Answer the following questions based on the full paper text:

### Question 1: Is there explicit validation?

**Does the paper explicitly compare LLM-as-a-judge results with human evaluation results?**

Answer "Yes" only if the paper:
- Compares LLM and human judgments on the same set of instances
- Reports quantitative metrics of agreement/correlation between LLM and human
- Discusses the relationship between LLM and human evaluation results

Answer "No" if:
- Both LLM and human evaluations are conducted but never compared
- LLM and human evaluate different sets of instances or different aspects
- Only qualitative discussion without any comparison

---

### Question 2: LLM-as-a-Judge Details

**A. Number of LLM Models**:
- How many different LLM models were used as judges?
- List the models (from Answer 3 metadata)

**B. LLM Prompts** (CRITICAL - Extract exact prompts if available):
- Does the paper show the exact prompt(s) used for LLM evaluation?
- If yes, extract the full prompt text verbatim
- If no, describe what information is provided about the prompts
- Note: Look for prompts in main text, tables, figures or appendices.

---

### Question 3: Human Evaluation Details

**A. Number of Human Evaluators**:
- How many human annotators/evaluators were used?

**B. Evaluator Type**:
- "expert": Domain experts, researchers, or trained annotators
- "crowdsourced": Crowd workers (MTurk, Prolific, etc.)
- "mixed": Combination of both
- "unclear": Not specified

**C. Inter-Annotator Agreement** (CRITICAL):
- Was inter-annotator agreement (IAA) reported?
- If yes, extract:
  - Metric used (Cohen's kappa, Fleiss' kappa, Krippendorff's alpha, percentage agreement, etc.)
  - Value(s) reported
  - Interpretation if provided (e.g., "substantial agreement")
- If no, note "Not reported"

**D. Human Evaluation Guidelines**:
- Does the paper provide detailed evaluation guidelines/instructions?
- Are example annotations or scoring rubrics shown?
- Where are guidelines described (main text, appendix, supplementary)?
---

### Question 4: Validation Setup (if Q1 = Yes)

**A. Validation Type** (select all that apply):
- "correlation_analysis": Correlation between LLM and human scores
- "agreement_analysis": Agreement between LLM and human labels/judgments
- "ranking_comparison": Compare rankings produced by LLM vs human
- "error_analysis": Analyze disagreements between LLM and human
- "other": Other types of validation

**B. Validation Metrics**:
Examples: "Pearson correlation", "Spearman correlation", "Kendall's tau", "Cohen's kappa", "Accuracy", "F1", "Krippendorff's alpha", etc.

**C. Shared Evaluation Criteria**:
List only criteria that both LLM and human evaluate.

**D. Sample Size**:
- How many instances/examples were used for validation?
- Note if validation uses subset or all evaluated instances

---

### Question 5: Validation Results (if Q1 = Yes)

**A. Correlation/Agreement Scores** (Extract ALL reported values):
For EACH metric reported, extract:
- Metric name (e.g., "Spearman correlation", "Cohen's kappa")
- Value (numerical - extract exact value as reported)
- Which criterion it applies to (e.g., "fluency", "coherence")
- Which LLM model (if multiple LLMs were compared)
- Statistical significance if reported (p-value, confidence intervals)

**B. Correlation Strength Interpretation**:
- Does the paper interpret correlation strength?
- What threshold do they use for "strong" correlation?
- Do they compare to prior work?

---

## Output Format

Return ONLY a JSON object with this structure:
{
  "paper_id": "{paper_id}",
  "explicit_validation": {
    "answer": "Yes"/"No",
    "explanation": "Brief explanation"
  },
  "llm_judge_details": {
    "num_models": 1,
    "models": ["GPT-4", ...],
    "prompts": {
      "provided": "yes"/"no"/"partial",
      "location": "appendix"/"main_text"/"figure"/"not_provided",
      "notes": "Additional notes"
    }
  },
  "human_eval_details": {
    "num_evaluators": 3,
    "evaluator_type": "expert"/"crowdsourced"/"mixed"/"unclear",
    "inter_annotator_agreement": {
      "reported": "yes"/"no",
      "metric": "Fleiss' kappa",
      "value": 0.68,
      "interpretation": "substantial agreement"
    },
    "guidelines": {
      "detailed_guidelines_provided": "yes"/"no"/"partial",
      "location": "appendix"/"main_text"/"not_provided"
    }
  },
  "validation_setup": {
    "validation_types": ["correlation_analysis", ...],
    "validation_metrics": ["Pearson correlation", ...],
    "shared_criteria": ["fluency", ...],
    "sample_size": {
      "total_generated": 100,
      "validated_by_both": 50
    }
  },
  "validation_results": {
    "quantitative_scores": [
      {
        "metric": "Spearman correlation",
        "value": 0.87,
        "criterion": "fluency",
        "llm_model": "GPT-4"
      }
    ],
    "summary_finding": "1-2 sentence summary"
  },
  "criteria_mapping": {
    "llm_only_criteria": [...],
    "human_only_criteria": [...],
    "shared_criteria": [...]
  }
}

## Important Notes

**If explicit_validation.answer = "No":**
- Set validation_setup and validation_results to `null`
- Still fill out llm_judge_details and human_eval_details
- Still fill out criteria_mapping

**Always extract (regardless of validation):**
- llm_judge_details: Always extract LLM setup information
- human_eval_details: Always extract human evaluation information
- criteria_mapping: Always show which criteria each method uses

**Guidelines:**
- **Be precise**: Only mark as validated if there's explicit comparison
- **Extract exact values**: Copy numerical results exactly as reported
- **Distinguish correlation types**: Pearson vs Spearman vs Kendall
- **Note statistical significance**: If p-values or confidence intervals are reported
- **Consider multi-criterion scenarios**: LLM and human might evaluate different criteria even in same paper

**Common Scenarios:**
1. **Full validation**: Paper uses LLM to evaluate all outputs, validates on human-annotated subset, reports correlation
2. **Parallel evaluation**: Both LLM and human evaluate the same outputs, direct comparison
3. **Sequential validation**: Human labels used as ground truth, LLM accuracy measured
4. **Independent streams**: Both methods used but never compared (answer "No")
5. **Qualitative only**: Paper discusses differences but no quantitative comparison (answer "No")

**Read the full paper text carefully** and extract all validation-related information accurately.
\end{tcblisting}

\label{append:prompt}

%
%

\tcbset{
    guidelinebox/.style={
        enhanced,
        width=\linewidth,
        colback=gray!5,
        colframe=black!50,
        boxrule=0.5pt,
        arc=2pt,
        breakable,
        pad at break*=2mm,
        left=5pt,
        right=5pt,
        top=5pt,
        bottom=5pt,
        before skip=0pt,
        after skip=0pt
    }
}

\section{Human Annotation Guideline}
\label{append:human}

To demonstrate, we show our human annotation guideline of the 110 manually annotated extractions below. For the guidelines of other types of annotations, please refer to our project repository.
\vspace{0.5em}

\begin{tcolorbox}[guidelinebox]
\noindent\textbf{Overview}

\vspace{0.5em}
\noindent This document provides detailed instructions for manually annotating research papers about natural language generation (NLG) evaluation. You will read each paper and extract structured metadata to answer four main questions about the paper's approach to NLG evaluation.

\vspace{0.5em}
\noindent\textbf{Important:} The papers you are annotating have been pre-filtered as potential NLG papers (with an initial ``Yes'' answer to Question 1). However, this filtering may not be perfect. You should verify this classification and change the answer to ``No'' if, after reading the paper, you determine it does not actually address NLG tasks.

\vspace{1em}
\noindent\textbf{Annotation Process}

\vspace{0.5em}
\noindent\textbf{Step 1: Read the Paper}

\begin{enumerate}[nosep,leftmargin=1.5em]
    \item Download and read the paper using the PDF link provided in the spreadsheet
    \item Take notes as you read to identify relevant information for each question
\end{enumerate}

\vspace{0.5em}
\noindent\textbf{Step 2: Answer Four Main Questions}

\vspace{0.3em}
\noindent For each paper, you will answer four yes/no questions and extract relevant metadata.

\vspace{1.5em}
\noindent\rule{\textwidth}{0.4pt}
\vspace{0.5em}

\noindent\textbf{Question 1: Does the paper address NLG tasks?}

\vspace{0.5em}
\noindent\textbf{Definition:}

\vspace{0.3em}
\noindent Natural Language Generation (NLG) refers to tasks where a system produces/generates natural language text as output. This is distinct from Natural Language Understanding (NLU) tasks where the system only reads/analyzes text.

\vspace{0.5em}
\noindent\textbf{Important Context:}

\vspace{0.3em}
\noindent These papers have been pre-filtered as NLG papers (initially classified as ``Yes''). However, the automatic filtering may have made mistakes. Your job is to verify this classification by carefully reading the paper.

\vspace{0.5em}
\noindent\textbf{How to Answer:}

\vspace{0.3em}
\noindent\textit{Answer ``Yes'' if:}
\begin{itemize}[nosep,leftmargin=1.5em]
    \item The paper generates text, sentences, or natural language as output
    \item The paper evaluates or studies systems that produce natural language
    \item Examples: summarization systems, dialogue systems, machine translation, paraphrase generation, image captioning
\end{itemize}

\vspace{0.3em}
\noindent\textit{Answer ``No'' if:}
\begin{itemize}[nosep,leftmargin=1.5em]
    \item The paper only does classification, tagging, or understanding tasks
    \item No text is generated as output
    \item Examples: sentiment analysis, named entity recognition, question answering with extractive answers (just selecting existing text)
    \item The paper was incorrectly classified during pre-filtering
\end{itemize}

\vspace{0.3em}
\noindent\textit{If you change the answer to ``No'':}
\begin{itemize}[nosep,leftmargin=1.5em]
    \item Explain your reasons for ``No'', for example, specify the tasks addressed in this paper
    \item Skip this paper entirely and move to the next paper
    \item You do not need to fill in any other fields (Q1 metadata, Q2, Q3, Q4)
\end{itemize}

\vspace{0.5em}
\noindent\textbf{Metadata to Extract (if answer is ``Yes''):}

\vspace{0.5em}
\noindent\textbf{1. Tasks (List of NLG task types)}

\vspace{0.3em}
\noindent\textit{What to include:}
\begin{itemize}[nosep,leftmargin=1.5em]
    \item The main NLG task(s) that the paper addresses
    \item Use standardized task names from this list:
    \begin{itemize}[nosep,leftmargin=2em]
        \item Text Summarization
        \item Dialogue Generation
        \item Paraphrase Generation
        \item Machine Translation
        \item Image Captioning
        \item Code Generation
    \end{itemize}
    \item If the task doesn't fit any category, use: ``Other: [specific task name]''
\end{itemize}

\vspace{0.3em}
\noindent\textit{Examples:}
\begin{itemize}[nosep,leftmargin=1.5em]
    \item \textcolor{green!60!black}{[GOOD]} ``Text Summarization'', ``Dialogue Generation''
    \item \textcolor{red}{[BAD]} Don't include: ``NLP'', ``Generation'' (too vague)
\end{itemize}

\vspace{0.3em}
\noindent\textit{Instructions:}
\begin{itemize}[nosep,leftmargin=1.5em]
    \item Include only the primary task(s) being studied/evaluated
    \item Don't include tasks mentioned only in related work or background
    \item If a paper studies multiple NLG tasks, list all of them
\end{itemize}

\vspace{0.5em}
\noindent\textbf{2. Datasets (List of dataset names)}

\vspace{0.3em}
\noindent\textit{What to include:}
\begin{itemize}[nosep,leftmargin=1.5em]
    \item Names of NLG datasets used for experiments or evaluation
    \item Include datasets that are central to the paper's contribution
    \item Use the official dataset name as cited in the paper
\end{itemize}

\vspace{0.3em}
\noindent\textit{Examples:}
\begin{itemize}[nosep,leftmargin=1.5em]
    \item \textcolor{green!60!black}{[GOOD]} ``CNN/DailyMail'', ``XSum'', ``WMT14'', ``MultiWOZ''
    \item \textcolor{red}{[BAD]} Don't include: Generic terms like ``news articles'', ``dialogue data''
\end{itemize}

\vspace{0.3em}
\noindent\textit{Instructions:}
\begin{itemize}[nosep,leftmargin=1.5em]
    \item Only include datasets that are actually used in the paper's experiments
    \item Don't include datasets only mentioned in related work
    \item If a paper creates a new dataset, include its name, if it is not named, use ``Proposed dataset for <task name>''
    \item Use the exact name from the paper
\end{itemize}

\vspace{0.5em}
\noindent\textbf{3. Languages (List of languages)}

\vspace{0.3em}
\noindent\textit{What to include:}
\begin{itemize}[nosep,leftmargin=1.5em]
    \item The language(s) of the generated outputs
    \item Use standard language names in English
\end{itemize}

\vspace{0.3em}
\noindent\textit{Examples:}
\begin{itemize}[nosep,leftmargin=1.5em]
    \item \textcolor{green!60!black}{[GOOD]} ``English'', ``Chinese'', ``German'', ``French''
    \item \textcolor{red}{[BAD]} Don't use: ISO codes like ``en'', ``zh'' (use full names)
\end{itemize}

\vspace{0.3em}
\noindent\textit{Instructions:}
\begin{itemize}[nosep,leftmargin=1.5em]
    \item Include all target languages for generation
    \item For multilingual papers, list all languages mentioned
    \item If the paper doesn't specify but uses English datasets, annotate as ``English''
\end{itemize}

\vspace{0.5em}
\noindent\textbf{4. Models (List of model names)}

\vspace{0.3em}
\noindent\textit{What to include:}
\begin{itemize}[nosep,leftmargin=1.5em]
    \item Names of NLG models used or proposed for GENERATION (not evaluation)
    \item These are models that produce/generate the text outputs
    \item Include both models proposed by the authors and baseline generation models
\end{itemize}

\vspace{0.3em}
\noindent\textit{IMPORTANT: This is for generation models only, NOT evaluation models:}
\begin{itemize}[nosep,leftmargin=1.5em]
    \item \textcolor{green!60!black}{[GOOD]} Include: Models that generate the summaries, translations, dialogue responses, etc.
    \item \textcolor{red}{[BAD]} Don't include: Models used to evaluate/judge outputs (those go in Q3)
    \item Example: If GPT-4 generates text $\rightarrow$ Q1. If GPT-4 judges/evaluates text $\rightarrow$ Q3.
\end{itemize}

\vspace{0.3em}
\noindent\textit{Examples:}
\begin{itemize}[nosep,leftmargin=1.5em]
    \item \textcolor{green!60!black}{[GOOD]} ``GPT-3'', ``BART'', ``T5'', ``Seq2Seq'', ``Transformer'' (if used for generation)
    \item \textcolor{green!60!black}{[GOOD]} ``GPT-3.5'', ``GPT-4'' (keep versions distinct when specified)
    \item \textcolor{red}{[BAD]} Don't include: Models only used for evaluation/judging outputs
\end{itemize}

\vspace{0.3em}
\noindent\textit{Normalization rules:}
\begin{itemize}[nosep,leftmargin=1.5em]
    \item Use canonical model names with proper capitalization
    \item Keep version numbers distinct: ``GPT-3'' vs ``GPT-4'' are different models
    \item Normalize case variations: ``gpt-3'' $\rightarrow$ ``GPT-3'', ``bert'' $\rightarrow$ ``BERT''
    \item Include size variants if specified: ``BERT-base'', ``BERT-large''
\end{itemize}

\vspace{0.3em}
\noindent\textit{Instructions:}
\begin{itemize}[nosep,leftmargin=1.5em]
    \item Focus on models that generate the NLG outputs being evaluated
    \item Don't list every model mentioned in passing in related work
    \item If a paper proposes a new generation model with a name, include it
    \item For papers proposing unnamed approaches, describe briefly: ``Proposed model with [backbone] architecture''
    \item Remember: Evaluation models go in Q3, not here!
\end{itemize}

\vspace{0.5em}
\noindent\textbf{5. Outputs (List of output descriptions)}

\vspace{0.3em}
\noindent\textit{What to include:}
\begin{itemize}[nosep,leftmargin=1.5em]
    \item Brief descriptions of what text is being generated
    \item Focus on the actual output artifacts, not the process
\end{itemize}

\vspace{0.3em}
\noindent\textit{Examples:}
\begin{itemize}[nosep,leftmargin=1.5em]
    \item \textcolor{green!60!black}{[GOOD]} ``News article summaries'', ``Task-oriented dialogue responses'', ``English-to-German translations'', ``Image captions''
    \item \textcolor{red}{[BAD]} Don't use long sentences: ``The approach generates task-oriented dialogue responses''
\end{itemize}

\vspace{0.3em}
\noindent\textit{Instructions:}
\begin{itemize}[nosep,leftmargin=1.5em]
    \item Use natural language descriptions
    \item Be specific but concise (3-7 words typically)
    \item If multiple types of outputs, list the main ones
    \item Focus on what is generated, not how
\end{itemize}

\vspace{1em}
\noindent\rule{\textwidth}{0.4pt}
\vspace{0.5em}

\noindent\textbf{Question 2: Does the paper use automatic metrics to evaluate the generated outputs?}

\vspace{0.5em}
\noindent\textbf{Definition:}

\vspace{0.3em}
\noindent Automatic metrics are computational measures that evaluate the quality of generated text without human involvement. These metrics compare generated text against reference texts or use logics, rules or learned models to score outputs.

\vspace{0.5em}
\noindent\textbf{How to Answer:}

\vspace{0.3em}
\noindent\textit{Answer ``Yes'' if:}
\begin{itemize}[nosep,leftmargin=1.5em]
    \item The paper reports scores from any automatic evaluation metrics
    \item Metrics are used to compare different systems or configurations
    \item Common examples: BLEU, ROUGE, METEOR, BERTScore, BLEURT, ChrF
\end{itemize}

\vspace{0.3em}
\noindent\textit{Answer ``No'' if:}
\begin{itemize}[nosep,leftmargin=1.5em]
    \item No evaluation of generated outputs is performed, or only uses human evaluation
\end{itemize}

\vspace{0.5em}
\noindent\textbf{Metadata to Extract (if answer is ``Yes''):}

\vspace{0.5em}
\noindent\textbf{Automatic Metrics (List of metric names)}

\vspace{0.3em}
\noindent\textit{What to include:}
\begin{itemize}[nosep,leftmargin=1.5em]
    \item All automatic evaluation metrics used to assess the generated outputs
    \item Use standardized metric names with proper capitalization, if unsure, check online sources
    \item A list of common evaluation metric names: \url{https://huggingface.co/evaluate-metric}
\end{itemize}

\vspace{0.3em}
\noindent\textit{Examples:}
\begin{itemize}[nosep,leftmargin=1.5em]
    \item \textcolor{green!60!black}{[GOOD]} ``BLEU'', ``ROUGE'', ``METEOR'', ``BERTScore'', ``BLEURT'', ``ChrF'', ``TER'', ``PARENT''
    \item \textcolor{red}{[BAD]} Don't use: ``bleu'', ``Blue'' (wrong capitalization)
\end{itemize}

\vspace{0.3em}
\noindent\textit{Critical normalization rule:}

\vspace{0.2em}
\noindent Simplify metric variants to their base form:
\begin{itemize}[nosep,leftmargin=1.5em]
    \item ``BLEU-1'', ``BLEU-2'', ``BLEU-4'' $\rightarrow$ all become ``BLEU''
    \item ``ROUGE-1'', ``ROUGE-2'', ``ROUGE-L'' $\rightarrow$ all become ``ROUGE''
    \item ``F1-score'', ``Exact Match'' $\rightarrow$ keep as separate metrics
    \item ``BERT-F1'', ``BERTScore'', ``Bertscore'' $\rightarrow$ use ``BERTScore''
\end{itemize}

\vspace{0.3em}
\noindent\textit{Why normalize variants?}
\begin{itemize}[nosep,leftmargin=1.5em]
    \item We want to know which metric families are used, not every variant
    \item This simplifies analysis and prevents overcounting similar metrics
\end{itemize}

\vspace{0.3em}
\noindent\textit{Instructions:}
\begin{itemize}[nosep,leftmargin=1.5em]
    \item Include all metrics actually used in the evaluation section
    \item Don't include metrics only mentioned in related work
    \item Use the base metric name (BLEU not BLEU-4)
\end{itemize}

\vspace{1em}
\noindent\rule{\textwidth}{0.4pt}
\vspace{0.5em}

\noindent\textbf{Question 3: Does the paper use Large Language Models (LLMs) as judges?}

\vspace{0.5em}
\noindent\textbf{Definition:}

\vspace{0.3em}
\noindent This refers to using LLMs after generation to automatically evaluate or judge the quality of generated outputs. The LLM is used as an evaluator, not as the generation model itself (except both use the same model).

\vspace{0.5em}
\noindent\textbf{How to Answer:}

\vspace{0.3em}
\noindent\textit{Answer ``Yes'' if:}
\begin{itemize}[nosep,leftmargin=1.5em]
    \item An LLM (like GPT-4, Claude, Llama) is used to score, rank, or judge generated outputs
    \item The paper describes using LLM prompts to assess quality
    \item Examples: ``GPT-4 as a judge'', ``LLM-based evaluation'', ``using ChatGPT to rate fluency''
\end{itemize}

\vspace{0.3em}
\noindent\textit{Answer ``No'' if:}
\begin{itemize}[nosep,leftmargin=1.5em]
    \item LLMs are only used for generation, not evaluation
    \item No LLM-based evaluation is performed
    \item Only traditional automatic metrics or human evaluation is used
\end{itemize}

\vspace{0.5em}
\noindent\textbf{Metadata to Extract (if answer is ``Yes''):}

\vspace{0.5em}
\noindent\textbf{1. Models (List of LLM names used as judges)}

\vspace{0.3em}
\noindent\textit{What to include:}
\begin{itemize}[nosep,leftmargin=1.5em]
    \item Names of specific LLMs used for evaluation
    \item Include version numbers when specified
\end{itemize}

\vspace{0.3em}
\noindent\textit{Examples:}
\begin{itemize}[nosep,leftmargin=1.5em]
    \item \textcolor{green!60!black}{[GOOD]} ``GPT-4'', ``GPT-3.5'', ``Claude-3'', ``PaLM-2'', ``Llama-2-70B''
    \item \textcolor{red}{[BAD]} Don't use: ``ChatGPT'' (use ``GPT-3.5-Turbo'' or ``GPT-4'' if version is known)
\end{itemize}

\vspace{0.3em}
\noindent\textit{Normalization rules:}
\begin{itemize}[nosep,leftmargin=1.5em]
    \item Use official model names with proper capitalization
    \item Keep versions distinct: ``GPT-3'' vs ``GPT-4''
    \item If paper says ``ChatGPT'' without version, keep as ``ChatGPT''
    \item Keep consistent format: ``ModelName-Version-Size'' (e.g., ``Claude-3-Opus'', ``Llama-2-70B'')
\end{itemize}

\vspace{0.5em}
\noindent\textbf{2. Methods (List of evaluation methods/approaches)}

\vspace{0.3em}
\noindent\textit{What to include:}
\begin{itemize}[nosep,leftmargin=1.5em]
    \item Brief description or name of the evaluation procedure
    \item How the LLM is prompted or used
\end{itemize}

\vspace{0.3em}
\noindent\textit{Examples:}
\begin{itemize}[nosep,leftmargin=1.5em]
    \item \textcolor{green!60!black}{[GOOD]} ``Pairwise comparison'', ``Direct scoring'', ``Likert scale rating'', ``Binary preference'', ``Multi-aspect scoring''
    \item \textcolor{green!60!black}{[GOOD]} ``Chain-of-thought evaluation'', ``Self-consistency''
    \item \textcolor{red}{[BAD]} Don't include: Full prompt text (too detailed)
\end{itemize}

\vspace{0.3em}
\noindent\textit{Instructions:}
\begin{itemize}[nosep,leftmargin=1.5em]
    \item Use short descriptive names (2-5 words)
    \item If the paper gives a name to their method, use it
    \item If not, describe the approach briefly
\end{itemize}

\vspace{0.5em}
\noindent\textbf{3. Criteria (List of evaluation criteria)}

\vspace{0.3em}
\noindent\textit{What to include:}
\begin{itemize}[nosep,leftmargin=1.5em]
    \item The specific aspects or dimensions that the LLM is asked to evaluate
    \item The rubric properties being scored
\end{itemize}

\vspace{0.3em}
\noindent\textit{Examples:}
\begin{itemize}[nosep,leftmargin=1.5em]
    \item \textcolor{green!60!black}{[GOOD]} ``Fluency'', ``Relevance'', ``Coherence'', ``Factuality'', ``Helpfulness'', ``Safety''
    \item \textcolor{red}{[BAD]} Don't include: The scores themselves (like ``1-5 scale'')
\end{itemize}

\vspace{0.3em}
\noindent\textit{If criteria are not specified:}
\begin{itemize}[nosep,leftmargin=1.5em]
    \item Leave the field empty if the paper doesn't mention specific criteria
    \item Don't guess or infer criteria
\end{itemize}

\vspace{0.3em}
\noindent\textit{Normalization rule:}
\begin{itemize}[nosep,leftmargin=1.5em]
    \item Use criteria names with proper capitalization: ``Fluency'' instead of ``fluency''
    \item Use nouns instead of adjectives: ``Naturalness'' instead of ``natural''
\end{itemize}

\vspace{0.3em}
\noindent\textit{Instructions:}
\begin{itemize}[nosep,leftmargin=1.5em]
    \item List all criteria mentioned
    \item Use the exact terminology from the paper when possible
    \item If paper uses very general terms like ``quality'', still include it
\end{itemize}

\vspace{1em}
\noindent\rule{\textwidth}{0.4pt}
\vspace{0.5em}

\noindent\textbf{Question 4: Does the paper conduct human evaluations of the generated outputs?}

\vspace{0.5em}
\noindent\textbf{Definition:}

\vspace{0.3em}
\noindent Human evaluation means that real people (not LLMs or automatic metrics) are asked to read and assess the generated outputs. This includes crowdsourcing, expert annotations, or user studies.

\vspace{0.5em}
\noindent\textbf{How to Answer:}

\vspace{0.3em}
\noindent\textit{Answer ``Yes'' if:}
\begin{itemize}[nosep,leftmargin=1.5em]
    \item Human annotators / raters evaluate the generated text
    \item User studies with human participants assess outputs after the generation, not for human annotated datasets used to training the generation model
\end{itemize}

\vspace{0.3em}
\noindent\textit{Answer ``No'' if:}
\begin{itemize}[nosep,leftmargin=1.5em]
    \item Only automatic metrics or LLM judges are used
    \item Humans are only used for data collection, not evaluation
\end{itemize}

\vspace{0.5em}
\noindent\textbf{Metadata to Extract (if answer is ``Yes''):}

\vspace{0.5em}
\noindent\textbf{1. Methods (List of evaluation methods/approaches)}

\vspace{0.3em}
\noindent\textit{What to include:}
\begin{itemize}[nosep,leftmargin=1.5em]
    \item Brief description or name of the evaluation procedure
    \item How human evaluators are asked to assess the outputs
    \item The type of evaluation task (rating, ranking, comparison, etc.)
\end{itemize}

\vspace{0.3em}
\noindent\textit{Examples:}
\begin{itemize}[nosep,leftmargin=1.5em]
    \item \textcolor{green!60!black}{[GOOD]} ``Pairwise comparison'', ``Direct scoring'', ``Likert scale rating'', ``Binary preference'', ``Multi-aspect rating''
    \item \textcolor{green!60!black}{[GOOD]} ``Ranking'', ``Best-worst scaling'', ``Magnitude estimation''
    \item \textcolor{green!60!black}{[GOOD]} ``A/B testing'', ``Adequacy and fluency rating''
    \item \textcolor{red}{[BAD]} Don't include: Full instruction text or specific questions (too detailed)
\end{itemize}

\vspace{0.3em}
\noindent\textit{Instructions:}
\begin{itemize}[nosep,leftmargin=1.5em]
    \item Use short descriptive names (2-5 words)
    \item If the paper gives a name to their evaluation method, use it
    \item If not, describe the approach briefly (e.g., ``5-point Likert scale rating'')
    \item If multiple different evaluation methods are used, list them separately
\end{itemize}

\vspace{0.5em}
\noindent\textbf{2. Criteria (List of evaluation criteria)}

\vspace{0.3em}
\noindent\textit{What to include:}
\begin{itemize}[nosep,leftmargin=1.5em]
    \item The specific aspects or dimensions that humans are asked to evaluate
    \item Evaluation categories or rubric items
\end{itemize}

\vspace{0.3em}
\noindent\textit{Examples:}
\begin{itemize}[nosep,leftmargin=1.5em]
    \item \textcolor{green!60!black}{[GOOD]} ``Fluency'', ``Adequacy'', ``Coherence'', ``Informativeness'', ``Naturalness'', ``Relevance'', ``Grammaticality'', ``Readability'', ``Factuality''
    \item \textcolor{red}{[BAD]} Don't include: The scores themselves
\end{itemize}

\vspace{0.3em}
\noindent\textit{If criteria are not specified:}
\begin{itemize}[nosep,leftmargin=1.5em]
    \item Leave it empty if the paper only describes a general ``quality'' rating without specific dimensions
    \item Don't infer criteria if not explicitly stated
\end{itemize}

\vspace{0.3em}
\noindent\textit{Normalization rule:}
\begin{itemize}[nosep,leftmargin=1.5em]
    \item Use criteria names with proper capitalization: ``Fluency'' instead of ``fluency''
    \item Use nouns instead of adjectives: ``Naturalness'' instead of ``natural''
\end{itemize}

\vspace{0.3em}
\noindent\textit{Instructions:}
\begin{itemize}[nosep,leftmargin=1.5em]
    \item Use exact terminology from the paper
    \item List all criteria mentioned in the evaluation setup
    \item If the paper uses ``overall quality'' as the only criterion, include it
\end{itemize}

\vspace{1em}
\noindent\rule{\textwidth}{0.4pt}
\vspace{0.5em}

\noindent\textbf{General Annotation Guidelines}

\vspace{0.5em}
\noindent\textbf{Quality Standards}

\begin{enumerate}[nosep,leftmargin=1.5em]
    \item \textbf{Be accurate:} Only annotate information that is explicitly stated in the paper
    \item \textbf{Be complete:} Try to find all relevant information for each question
    \item \textbf{Be consistent:} Use standardized names and formats
    \item \textbf{Be conservative:} When in doubt, don't guess---leave it out or mark as uncertain
\end{enumerate}

\vspace{0.5em}
\noindent\textbf{Handling Edge Cases}

\vspace{0.3em}
\noindent\textit{If you're unsure about something:}
\begin{itemize}[nosep,leftmargin=1.5em]
    \item Add a comment/note in your annotation
    \item Mark items you're uncertain about
    \item It's better to be cautious than incorrect
\end{itemize}

\vspace{0.3em}
\noindent\textit{If information is ambiguous:}
\begin{itemize}[nosep,leftmargin=1.5em]
    \item Use your best judgment based on context
    \item Add a note explaining your interpretation
\end{itemize}

\vspace{0.3em}
\noindent\textit{If a field should be empty:}
\begin{itemize}[nosep,leftmargin=1.5em]
    \item Leave it empty, don't put placeholder text
\end{itemize}

\vspace{0.5em}
\noindent\textbf{Normalization Standards}

\vspace{0.3em}
\noindent\textit{Capitalization:}
\begin{itemize}[nosep,leftmargin=1.5em]
    \item Metrics: Follow standard conventions (BLEU, ROUGE, BERTScore)
    \item Models: Use official capitalization (GPT-4, BERT, T5)
    \item Tasks: Use title case (Text Summarization, Machine Translation)
    \item Languages: Capitalize (English, Chinese, German)
\end{itemize}

\vspace{0.3em}
\noindent\textit{Naming:}
\begin{itemize}[nosep,leftmargin=1.5em]
    \item Use official, canonical names when available
    \item Be consistent across all annotations
    \item Merge obvious duplicates (e.g., ``MT'' and ``Machine Translation'' $\rightarrow$ use ``Machine Translation'')
\end{itemize}

\vspace{0.5em}
\noindent\textbf{Common Mistakes to Avoid}

\vspace{0.3em}
\noindent\textcolor{red}{\textbf{[DON'T INCLUDE]:}}
\begin{itemize}[nosep,leftmargin=1.5em]
    \item Information from related work sections (unless actually used in the paper)
    \item Background or motivation content (focus on what the paper does)
    \item Every model/dataset mentioned (focus on what's evaluated)
\end{itemize}

\vspace{0.3em}
\noindent\textcolor{green!60!black}{\textbf{[DO INCLUDE]:}}
\begin{itemize}[nosep,leftmargin=1.5em]
    \item Information from experiments and evaluation sections
    \item Main contributions and findings
    \item All metrics, criteria, and methods actually used
    \item Clear, specific terminology from the paper
\end{itemize}

\vspace{1em}
\noindent\rule{\textwidth}{0.4pt}
\vspace{0.5em}

\noindent\textbf{Annotation Workflow Summary}

\vspace{0.3em}
\noindent\textbf{For each paper:}

\begin{enumerate}[nosep,leftmargin=1.5em]
    \item Read the paper (especially abstract, methodology, and evaluation sections)
    \item Answer Question 1: Does it address NLG tasks?
    \begin{itemize}[nosep,leftmargin=2em]
        \item Verify the pre-filtered classification - the paper was initially classified as ``Yes''
        \item Change to ``No'' if it doesn't actually address NLG tasks
        \item If No: Skip to the next paper (we only annotate NLG papers)
        \item If Yes: Continue to extract all metadata
    \end{itemize}
    \item Extract Q1 metadata: tasks, datasets, languages, models, outputs
    \item Answer Question 2: Does it use automatic metrics?
    \begin{itemize}[nosep,leftmargin=2em]
        \item If Yes: Extract and normalize metric names
    \end{itemize}
    \item Answer Question 3: Does it use LLMs as judges?
    \begin{itemize}[nosep,leftmargin=2em]
        \item If Yes: Extract LLM models, methods, and criteria
    \end{itemize}
    \item Answer Question 4: Does it conduct human evaluation?
    \begin{itemize}[nosep,leftmargin=2em]
        \item If Yes: Extract evaluation methods and criteria
    \end{itemize}
    \item Review your annotations for completeness and consistency
    \item Add any notes about difficult decisions or uncertainties
\end{enumerate}

\vspace{1em}
\noindent\textbf{Questions or Issues?}

\vspace{0.3em}
\noindent If you encounter any problems during annotation:
\begin{itemize}[nosep,leftmargin=1.5em]
    \item Document unclear cases in the notes section
    \item Flag papers that are ambiguous or difficult to categorize
    \item Ask for clarification on systematic issues
\end{itemize}

\end{tcolorbox}

\end{document}